\documentclass[lettersize,journal]{IEEEtran}
\usepackage{amsmath,amsfonts}
\usepackage{algorithm}
\usepackage{algorithmic}
\usepackage{array}
\usepackage[caption=false,font=normalsize,labelfont=sf,textfont=sf]{subfig}
\usepackage{threeparttable}
\usepackage{textcomp}
\usepackage[dvipsnames]{xcolor}
\usepackage{stfloats}
\usepackage{url}
\usepackage{verbatim}
\usepackage{graphicx}
\usepackage{cite}
\usepackage{multirow}
\usepackage[inkscapeformat=png]{svg}
\usepackage[colorlinks=true,linkcolor=NavyBlue,citecolor=BrickRed]{hyperref}
\usepackage[capitalise]{cleveref}
\usepackage[normalem]{ulem}
\usepackage[switch]{lineno}
\definecolor{amber}{rgb}{1.0, 0.75, 0.0}

\newcolumntype{?}{!{\vrule width 1.2pt}}

\begin{document}

\title{\textit{AIO2}: Online Correction of Object Labels for Deep Learning with Incomplete Annotation in Remote Sensing Image Segmentation}

\author{Chenying Liu,~\IEEEmembership{Student Member,~IEEE}, 
Conrad M Albrecht,~\IEEEmembership{Member,~IEEE}, 
Yi Wang,~\IEEEmembership{Student Member,~IEEE}, 
Qingyu Li, 
Xiao Xiang Zhu,~\IEEEmembership{Fellow,~IEEE}
        % <-this % stops a space
\thanks{C. Liu (chenying.liu@dlr.de) and Y. Wang (Yi.Wang@dlr.de) are with the Chair of Data Science in Earth Observation, Technical University of Munich (TUM), and the Remote Sensing Technology Institute, German Aerospace Center (DLR). C. M. Albrecht (Conrad.Albrecht@dlr.de) is with the Remote Sensing Technology Institute, German Aerospace Center (DLR). Q. Li (qingyu.li@tum.de) and X. X. Zhu (xiaoxiang.zhu@tum.de) are with the Chair of Data Science in Earth Observation, Technical University of Munich (TUM).}
}

% The paper headers
\markboth{Accepted by IEEE Transactions on Geoscience and Remote Sensing}%
{Shell \MakeLowercase{\textit{et al.}}: A Sample Article Using IEEEtran.cls for IEEE Journals}

\IEEEpubid{0000--0000/00\$00.00~\copyright~2021 IEEE}
% Remember, if you use this you must call \IEEEpubidadjcol in the second
% column for its text to clear the IEEEpubid mark.

\maketitle

\begin{abstract}
While the volume of remote sensing data is increasing daily, deep learning in Earth Observation faces lack of accurate annotations for supervised optimization. Crowdsourcing projects such as OpenStreetMap distribute the annotation load to their community. However, such annotation inevitably generates noise due to insufficient control of the label quality, lack of annotators, frequent changes of the Earth's surface as a result of natural disasters and urban development, among many other factors.\\
We present \textit{Adaptively trIggered Online Object-wise correction (AIO2)} to address annotation noise induced by incomplete label sets. AIO2 features an \textit{Adaptive Correction Trigger (ACT)} module that avoids label correction when the model training under- or overfits, and an \textit{Online Object-wise Correction (O2C)} methodology that employs spatial information for automated label modification. AIO2 utilizes a mean teacher model to enhance training robustness with noisy labels to both stabilize the training accuracy curve for fitting in ACT and provide pseudo labels for correction in O2C. Moreover, O2C is implemented \textit{online} without the need to store updated labels every training epoch. We validate our approach on two building footprint segmentation datasets with different spatial resolutions. Experimental results with varying degrees of building label noise demonstrate the robustness of AIO2. Source code will be available at \url{https://github.com/zhu-xlab/AIO2.git}.
\end{abstract}

\begin{IEEEkeywords}
Building detection, curriculum learning, deep learning, early learning, label correction, memorization effects, noisy labels, remote sensing, semantic segmentation.
\end{IEEEkeywords}

\section{Introduction} \label{sec:intro}
% 1 - lack of labels
\IEEEPARstart{D}{eep} learning has become a powerful tool of big data mining in Earth Observation (EO) \cite{zhu2017DLReview}. 
However, supervised deep learning methods are notorious data-hungry, requiring large amounts of high-quality labeled data to avoid overfitting. Despite the abundance of Remote Sensing (RS) images, obtaining accurately annotated labels poses a significant challenge due to the expensive, laborious, and time-consuming nature of the annotation process, which often involves domain experts and field surveys. 

% 2 - label noise resources
Nevertheless, there are many sources of labels from which we can easily obtain large amounts of labeled data with minimal efforts. For instance, Volunteered Geographic Information sources like OpenStreetMap (OSM) collect label information from individuals in a volunteer capacity and make it freely available \cite{Vargas2021OSM}. Another approach is to design automatic labeling tools, such as \textit{AutoGeoLabel} \cite{albrecht2021autogeolabel}, to generate labels rapidly for RS images from high-quality data sources e.g., LiDAR (Light Detection and Ranging) data. Additionally, various land use land cover products, including Google's Dynamic World \cite{brown2022DynamicWorld}, ESA's World Cover \cite{Zanaga2021WorldCover}, and Esri's Land Cover \cite{Karra2021ESRI}, offer rich information for EO. Nevertheless, these label sources often result in unreliable labels, {e.g., noisy labels due to insufficient human annotation. For example, \cite{zhu2020so2sat} documents human uncertainty for the classification of local climate zones.} As reported in \cite{zhang2021understanding}, deep learning models are known for their large number of parameters and capability of learning complex functions, yet vulnerability to label noise. This also applies to segmentation tasks \cite{liu2022peas}. Therefore, these readily available labels require special considerations when applied to real-world scenarios. Beyond model training, noisy labels may significantly affect the evaluation of methodologies as well \cite{hansch2019truth}.

\IEEEpubidadjcol

% 3 - learning from noisy labels, targeting on segmentation tasks, point out problems 
% borrow ideas from classification and semi-supervised learning domains 
% 1> from classification; 
While learning from noisy labels (LNL) has been extensively studied for image classification tasks, few approaches have been developed for image segmentation tasks. Existing LNL methods for segmentation tasks mainly borrow ideas from LNL for classification and semi-supervised segmentation methods. In the former case from classification tasks, a set of regularization techniques such as consistency regularization \cite{wang2020consist} or entropy minimization \cite{li2022road} is used to constrain the optimization space. Nevertheless, label noise behaves differently in these two types of tasks. In classification tasks, the entire image is treated as a single sample unit and can be considered to have approximately similar levels of uncertainty. Thus, random flipping can be used to simulate label noise for classification tasks. In contrast, the sample unit in segmentation tasks is a pixel, and neighboring pixels are interconnected through spatial dependencies \cite{he2018hsireview}. As a result, pixels located near boundaries are more difficult to define. From this perspective, we can classify pixel-wise label noise into two categories: assignment noise and shape noise. Assignment noise occurs when objects are labeled incorrectly, while shape noise refers to inexact object delineation caused by such phenomena as coarse annotations. {In practice, inaccurate co-registration of image-mask pairs is another common source of label noise, mainly leading to shape noise with misaligned boundaries \cite{Maiti2022misalign}.} Generally, assignment noise incurs more severe damage on model training than shape noise does. {This difference is illustrated in \Cref{sec:meth:memo}.}
{Moreover, LNL for natural image segmentation is usually studied in the context of weakly supervised learning, where pixel-wise noisy labels are derived with image-level annotations by GradCAM and its variants from object-centric images \cite{shen2023survey, zhang2020reliability}. Thus, the primary label noise is the shape noise, while RS applications usually face more complex noise types due to different image characteristics and more diverse noisy label sources.}

% 2> from semi-supervised learning
As regards the ideas borrowed from the semi-supervised learning domain, self-training methods are naturally related to the noisy label problem, where pseudo labels generated by the classifier itself inevitably incur some inaccurate assignments {\cite{zhang2023selftrain}}. Following this paradigm, \cite{cao2022green, dong2022landcover, cao2023building} correct possibly wrong labels in the training set by means of high-confidence or low-uncertainty predictions. To make these methods effective, the questions of when and how to correct the labels should be considered.
In semi-supervised scenarios, only a small part of the accurately labeled patches is available at the beginning. The training set is gradually expanded via adding pseudo labels as training continues, during which the impact of bad pseudo labels can be offset to some extent by the advantages brought by training size expansion. LNL settings do not confer this advantage since the classifier originally has access to a large number of labels that are not as accurate as expected. Therefore, manually setting the warm-up length as in \cite{dong2022landcover} can easily lead to an early or late start of correction, risking correction effectiveness degradation when model predictions are not reliable enough. Liu et al. \cite{liu2022adele} propose an adaptive method for label correction initialization in which the training accuracy curve is fit on an exponential function and the change in its gradients is monitored. While promising, this method has a sensitive threshold setting to noise rates, and the fluctuation of accuracy curves makes the detection results unstable. In terms of how to correct, current correction criteria usually take softmax/sigmoid outputs as confidence indicators \cite{cao2023building}. The threshold is either predefined by users or flexibly adjusted in an image-wise fashion \cite{dong2022landcover}, more or less ignoring the spatial dependencies among pixels. {One further possibility is to determine data and model uncertainty, e.g.,\ via Bayesian Neural Networks, for sample selection \cite{gawlikowski2023uncertianty}. Yet a major challenge is the lack of ground truth to evaluate the estimated data uncertainty when developing such methods to address real-world problems.}

% Settings
In this work, we study building footprint identification from aerial imagery to develop a novel methodology to handle, among other types of noise, incomplete label noise, {in which a given set of building outlines is known to miss a subset of existing buildings (false negative), but annotated buildings are assumed to have accurate outlines. Existing research on quality assessment of OSM building data has found that in most areas position accuracy is comparable to cadastral maps, while completeness is relatively low, and that it varies worldwide \cite{brovelli2018osmbuildingassess, zhang2022osmbuildingassess, herfort2023buildingcomplete}. As mentioned above, assignment noise potentially imposes a more significant negative impact on model training than shape noise. Thus, we focus our study on incomplete label noise as a first step towards a systematic solution to using OSM labels for model training. 
Our approach significantly differs from semi-supervised scenarios, where all the labeled patches are carefully annotated, with all the objects marked. For us, all the patches are labeled with objects dropped from the ground truth by annotation as background. Thus, we call it ``incomplete label noise.''}
% We don't pay too much attention to the shape noise as 1) OSM has its own quality control of added annotations, and 2) slight deformation of shapes won't

% method
Based on the aforementioned issues, we propose a new method called Adaptively trIggered Online Object-wise correction (AIO2). This approach consists of two main components, an Adaptive Correction Trigger (ACT) module and an Online Object-wise label Correction (O2C) module, to address the ``when'' and ``how'' questions in the self-cleansing process without human interference. In short, AIO2 adopts ACT to automatically trigger the O2C by monitoring the dynamics of training accuracy curves measured by numerical gradients. Specifically, our framework incorporates a mean teacher model \cite{tarvainen2017meanteacher} originally designed for semi-supervised learning. The teacher model is updated by exponentially averaging historical model weights, thus leading to a minimal extra computational burden without backpropagation. In turn, the training accuracy curves by the teacher model are smoother, which can reduce the negative effects of fluctuations on early learning detection results. Moreover, we partially decouple the online label correction process and the model training by utilizing the predictions from teacher models as pseudo labels, with which we design an object-wise correction module for label cleansing. The main contributions of our work are as follows: 
\begin{enumerate}
    \item {We introduce a new label correction method termed AIO2 for segmentation tasks with incomplete label noise, which is less sensitive to parameter settings and more compatible with spatial characteristics of pixels.}
    \item {We analyze in detail the memorization effects in segmentation tasks as a basis for our methodology design. The resulting insights have served as valuable input for future extensions of noisy label training programs.}
    \item {We present two new modules, namely, ACT and O2C, which are particularly designed for segmentation tasks to solve the ``when" and ``how" problems in the self-cleansing process without human interference.}
   % an ACT module based on numerical gradient estimation to adaptively trigger label correction without manually setting the length of the warm-up stage, as well as an object-wise label correction strategy as a substitute for pixel-wised label correction guided by model behaviors in segmentation tasks.} 
\end{enumerate}

{In a nutshell, our methodology exploits the spatial context of pixels to explore memorization effects in pixel-level segmentation tasks. The high-level objective of semantic segmentation from remote sensing modalities is the generation of map data. \textit{Vectorizing} rasterized segmentation maps involves grouping pixels into single identities such as buildings, and other geospatial ``objects.'' Our work devises strategies for the analysis and adjustment of geospatial image semantic segmentation tasks at the object level, such as evaluation of training performance, label correction, and uncertainty assignment of pixels based on relative position (object ``boundary'' vs.\ ``bulk'').} 

The article is organized as follows: \Cref{sec:relate_works} summarizes related studies on LNL with deep learning models. Next, the memorization effects of noisy labels in segmentation tasks along with technical details of the proposed AIO2 method are described in \cref{sec:meth}. We elaborate the experimental results in \cref{sec:exp}, and conclude this work with some discussions and future lines in \cref{sec:conclude}.

\section{Related works} \label{sec:relate_works}

In this section, we review some recent advances in LNL with deep learning models for image classification and segmentation tasks, with a special focus on the RS domain.

\subsection{LNL for Image Classification Tasks}

The problem of noisy labels is well-investigated in the image classification field. One promising label source is web crawling: it is easy and cheap to obtain a large amount of labeled data, although it is somewhat unreliable \cite{wei2022learning}. 
To reduce the negative effects of label noise on model training, existing studies seek to find solutions using three main approaches: robust architecture modification \cite{sukhbaatar2014training,bekker2016training,goldberger2017training}, label cleansing \cite{tanaka2018joint, pleiss2020identifying, malach2017decoupling, jiang2018mentornet, han2018coteaching, yu2019coteachingplus, li2020dividemix, zhang2022ideal}, and robust loss function design \cite{tarvainen2017meanteacher, liu2020elr, yi2019pencil, Reed2015boot, zhang2018generalized, lyu2019curriculum, liu2020peer}. A comprehensive review of LNL methods for classification is provided in \cite{song2022learning}. Each of the three methods mentioned above is described briefly below.

A key concept of robust architecture modification is to add a label transition layer on top of a softmax layer of base deep neural networks in order to explicitly transfer the hidden ``true'' labels to their noisy versions for training \cite{sukhbaatar2014training}. In the test phase, the transition layer is removed to enclose ``clean" predictions. It can be modeled in a feature-independent fashion \cite{bekker2016training} or a feature-conditional way \cite{goldberger2017training}. Label cleansing is more straightforward than the other two method types, employing sample selection or correction. Incorrectly assigned labels can be recognized according to high uncertainty quantified by loss \cite{tanaka2018joint} or softmax outputs \cite{pleiss2020identifying}. Some works also leverage the discrepancy between simultaneously trained deep neural networks such as Decouple \cite{malach2017decoupling}, MentorNet \cite{jiang2018mentornet}, Co-teaching \cite{han2018coteaching, yu2019coteachingplus}, and DivideMix \cite{li2020dividemix}. Another reframes the noisy label problem as a domain shift one, and separates clean and noisy labels with the aid of data augmentation \cite{zhang2022ideal}. Robust loss function design is probably the most popular of the LNL methods, partially due to its flexibility and {its theoretical basis in risk minimization} \cite{manwani2013noise, ghosh2017robust}. 
%It can work solely \textcolor{red}{[]} or be combined with e.g., sample correction to \textcolor{red}{[]}. 
Some state-of-the-art methods include a consistency constraint between teacher and student model predictions in a self-ensembling framework \cite{tarvainen2017meanteacher}, early-learning regularization by integrating historical predictions to combat the memorization phenomenon \cite{liu2020elr}, compatibility loss between corrected label distributions and original noisy labels to avoid crazy deviation of corrections from original labels \cite{yi2019pencil}, bootstrapping using the convex combination of original noisy labels and predictions to prevent direct fitting on the noise distribution \cite{Reed2015boot}, and so on \cite{zhang2018generalized, lyu2019curriculum, liu2020peer}.

Within the RS community, the typical type of image classification task is scene classification, in which the noisy label problem is studied from the single-label and multi-label aspects. In single-label cases, some ideas are borrowed directly from the computer vision domain, such as using co-teaching for sample selection \cite{tai2020coastal}, and smooth loss to constrain the optimization process \cite{huang2020smoothloss}. Specifically for RS data, Damodaran et al. \cite{damodaran2020entropic} propose an entropic optimal transport loss inherently exploiting the geometric structure of the underlying data. Other methods are designed from the feature learning perspective, either doing sample cleansing \cite{li2020error,tu2020robust} or modifying the loss function to be noise-robust \cite{kang2021neighbor, kang2020rnsl}. In terms of multi-class scene classification with noisy labels, only a few works address the problem mainly employing loss design \cite{burgert2022multi, Aksoy2022multi} and sample selection \cite{Sumbul2023multilabelgenerative}.

\subsection{LNL for Image Segmentation Tasks}

{Unlike the extensive research in image classification, related LNL works for segmentation are relatively rare in the computer vision domain, which is generally studied along with weakly supervised methods using pixel-wise labels derived from activation maps guided by image-level annotations as noisy labels \cite{shen2023survey, zhang2020reliability}. 
These labels for object-centric images are primarily contaminated by shape noise. 
On the contrary, noisy labels are more frequently encountered in RS image segmentation tasks, where pixel-wise annotations are more difficult and ambiguous in combination with the clustered background, and thus require more expertise. 
Furthermore, there are more noisy label sources for RS image segmentation, such as OSM and various land use land cover products.} To combat noisy labels, some works have been designed under the assumption that a small number of clean labels are available during training \cite{maggiori2016convolutional, kaiser2017learning, ahmed2021dense}. However, this is not the case in many real scenarios. We thus concentrate on methods without usage of clean labels in the following review.

Inspired by LNL methods for classification, \cite{li2020tranlayer} and \cite{Zhang202tranlayer} appended a probabilistic model to ordinary deep neural networks in order to capture the relationship between noisy labels and their latent true counterparts for road and building extraction. Nevertheless, more common and effective solutions employ robust loss functions such as bootstrapping \cite{Henry2021boot}, consistency constraints \cite{malkin2019superreso, wang2020consist}, and loss reweighting with weights of each sample estimated by an attention mechanism \cite{lin2021atten} or reliability \cite{zhang2020reliability}. These methods, though effective to some extent, are sensitive to parameter setting, and sometimes unable to generalize well due to the long-tail distribution problem in segmentation tasks. In addition, Albrecht et al.\ \cite{albrecht2020change} exploit a CycleGAN for iterative addition of missing labels from style-translation of aerial images into rasterized OSM scenes. This approach incorporates spatial correlations and geographic context of human infrastructure such as roads, buildings, and parks.
On the other hand, encouraged by the connection of the noisy label problem with semi-supervised learning, confidence/uncertainty-based pixel-wise sample selection or correction is widely used, in which model performance is highly dependent on the predefined threshold setting \cite{cao2023building}. To alleviate this drawback, Dong et al. \cite{dong2022landcover} involve a patch-based threshold adjustment technique that can partially release the dependency on manual threshold setting. They also combine it with regularization constraining on original noisy labels, a popular strategy aiming to reduce the negative effects of mistaken corrected labels \cite{cao2022green}. 
{Besides, Sun et al. utilized mutual teaching with two structurally identical models to update noisy pseudo labels for hyperspectral image change detection \cite{sun2022hsi}.}
{The aforementioned approaches for RS image segmentation with noisy labels primarily rely on pixel-wise correction, with some adjustments to enhance adaptability to RS images. However, this pixel-wise constraint neglects crucial spatial information shared by neighboring pixels, a key factor in segmentation tasks. Additionally, these methods require a manually set warm-up stage, introducing instability. In a recent work, Liu et al. \cite{liu2022adele} proposed an adaptive early learning detection for medical image segmentation with noisy labels. While promising, its application to RS images proved unstable due to sensitive hyperparameter settings and susceptibility to accuracy curve fluctuations. In response, our proposed method, AIO2, is developed to achieve more robust early learning detection and enhance the effectiveness of sample correction with spatial information.}

\section{Methodology} \label{sec:meth}

\begin{figure*}
    \centering
    \includegraphics[width=2\columnwidth]{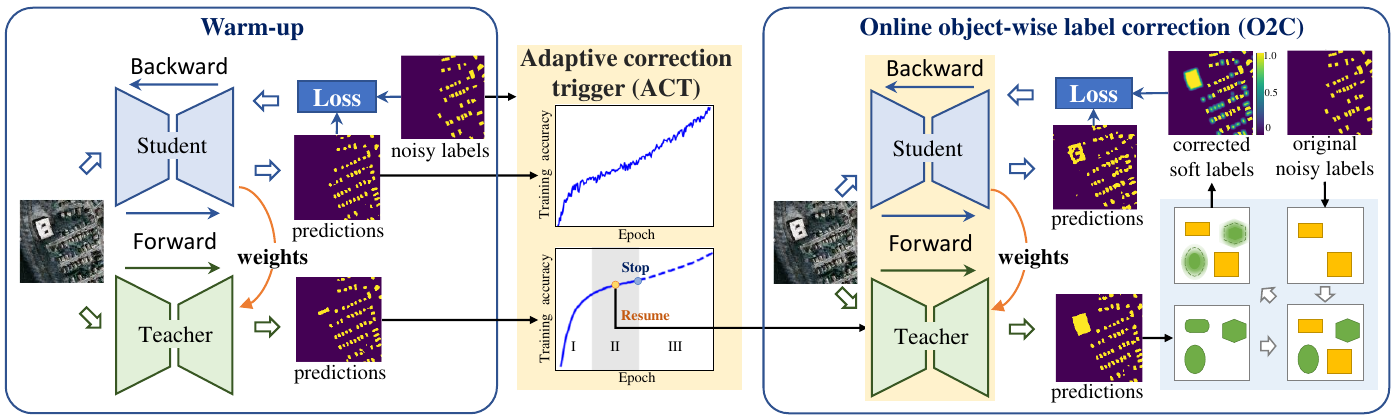}
    \caption{{\textit{Flowchart of the proposed two-stage AIO2 method for object-level incomplete label sets}: Model training is initially conducted using the given noisy labels, where ACT actively monitors the training dynamics to determine when to trigger O2C for label correction.
    %Flowchart of the proposed AIO2 method for object-level incomplete label sets}: {AIO2 is mainly composed of two modules, the Adaptive Correction Trigger (ACT) module and the Online Object-wise label Correction (O2C) module.} In addition, a teacher model is introduced whose weights are updated by exponential moving average (EMA) on historical weights of the student model. The teacher model on the one hand provides more smooth training accuracy curves for the ACT module to automatically terminate the warm-up phase and simultaneously trigger O2C for label cleansing, and on the other hand serves as the pseudo label source in O2C, and thus is able to partly decouple the label correction process from model training.
    }}
    \label{fig:flowchart}
\end{figure*}

\begin{figure*}[htp]
    \centering
    \begin{tabular}{m{0.45\linewidth}<{\centering}m{0.45\linewidth}<{\centering}}
         \includegraphics[width=0.75\linewidth]{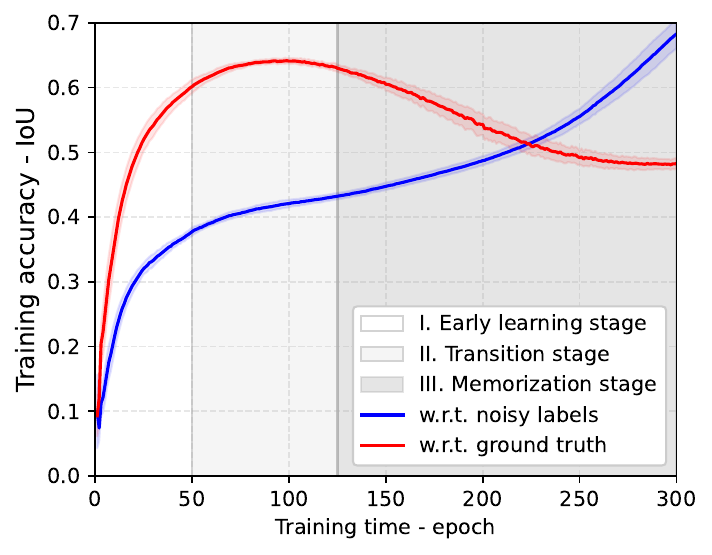} &  
         \includegraphics[width=0.78\linewidth]{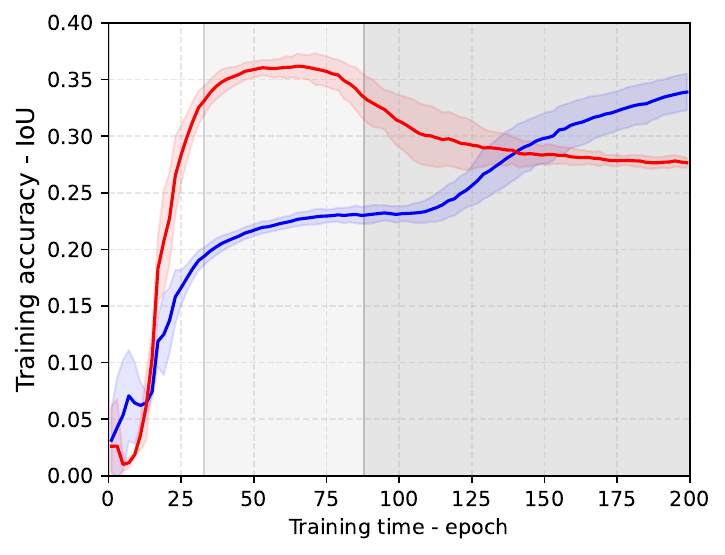} \\
         (a) Massachusetts dataset & (b) Germany dataset
    \end{tabular}
    \caption{{\textit{Three-stage training without special considerations for label noise (colors online)}:
    training accuracies of teacher models obtained with incomplete noisy labels of a drop rate of 0.5 on the (a) Massachusetts dataset and the (b) Germany dataset.
    Note: For real-world scenarios, training accuracies (\textcolor{blue}{blue}) needs to be based on noisy labels. Ground-truth label accuracies (\textcolor{red}{red}) are presented for reference only.}
    \newline For this figure and all those that follow, statistically fluctuating accuracy curves have been smoothed, with the solid line indicating the mean value and the shaded, semi-transparent region marking the $1\sigma$-area.}
    \label{fig:meth:memo3stages}
\end{figure*}

\Cref{fig:flowchart} presents an overview of the proposed AIO2 method. Analogous to other correction-based methods, AIO2 is initialized from a warm-up stage, taking given noisy labels as reference data to train the network. The Adaptive Correction Trigger (ACT) module at the same time ceases the training when the model starts overfitting to noisy labels. Both the student and teacher models are then reloaded from a previous checkpoint according to the refined detection result. Thereafter, training is resumed with Online Object-wise label Correction (O2C) coming into force. 
{
In this procedure, a teacher model is introduced whose weights are updated by exponential moving average (EMA) on historical weights of the student model. The teacher model on the one hand provides more smooth training accuracy curves for the ACT module to automatically terminate the warm-up phase and simultaneously trigger O2C for label cleansing, and on the other hand provide pseudo labels for O2C, and thus is able to partly decouple the label correction process from model training.
}
In the following, \cref{sec:meth:teacher} first gives a brief description of the mean teacher model. Some insights on memorization effects are discussed in \cref{sec:meth:memo}, followed by technical details of the ACT and O2C modules in \cref{sec:meth:act} and \cref{sec:meth:o2c}, respectively.

\subsection{Mean Teacher Model}
\label{sec:meth:teacher}

Temporal ensembling was first introduced into semi-supervised learning domain by implementing an exponential moving average (EMA) of historical successive predictions on each training example \cite{laine2017selfensemble}. Encouraged by its success, mean teacher modeling was developed by applying EMA on model weights instead of predictions. The effectiveness of mean teacher modeling to combat label noise was evaluated on classification tasks \cite{tarvainen2017meanteacher}. Doing so conferred two obvious advantages: better model stability, and a decreased storage and computational burden. In our work, we found that the stability of the mean teacher model is beneficial to both of our newly designed modules.

Let $\mathbf{\theta}_n^{(s)}$ denote the parameters of the student model at the $n$-th iteration updated in a regular model training approach through backpropagation:
\begin{equation} \label{eq:teacher:sup}
    \mathbf{\theta}_n^{(s)} = \mathbf{\theta}^{(s)}_{n-1} -\eta\nabla\mathcal{L}(\mathbf{\theta}^{(s)}_{n-1}),
\end{equation}
with $\eta$ as the learning rate, and $\nabla\mathcal{L}(\cdot)$ the gradients of loss function wrt each parameter. After the update of $\mathbf{\theta}_s^n$, the counterpart of teacher model $\mathbf{\theta}_t^n$ can be derived via EMA by
\begin{equation} \label{eq:teacher:tup}
    \mathbf{\theta}_n^{(t)} =
    \begin{cases}
      \mathbf{\theta}_{n}^{(s)} & n=0 \\
      \alpha\mathbf{\theta}_{n-1}^{(t)} +(1-\alpha)\mathbf{\theta}_{n}^{(s)} & n>0,
    \end{cases}
\end{equation}
where $\mathbf{\theta}_{0}^{(s)}$ are the randomly initialized models, and $\alpha$ is the smoothing coefficient hyperparameter empirically set as 0.999 \cite{tarvainen2017meanteacher}. 

\subsection{Memorization Effects} \label{sec:meth:memo}
Memorization effects were first reported on image classification tasks \cite{zhang2021understanding, arpit2017closer}, implying a two-stage training with noisy labels. More precisely, in the first \textit{early-learning} stage, model performance is continuously improved by dominant learning from most of the accurately labeled samples, while in the later \textit{memorization} stage, model performance begins to be degraded for overfitting to label noise information. A similar phenomenon has also been observed in segmentation tasks \cite{liu2022adele,liu2022peas}. Unlike the original elucidation of memorization effects, we re-interpret this phenomenon as a three-stage training with noisy labels, adding a transition stage between the early-learning and memorization stages (see \cref{fig:meth:memo3stages}). In the following we inspect this phenomenon at both pixel and object levels.

\begin{figure*}
    \centering
    \begin{tabular}{m{0.18\linewidth}<{\centering}m{0.23\linewidth}<{\centering}m{0.23\linewidth}<{\centering}m{0.23\linewidth}<{\centering}}
        \includegraphics[width=0.98\linewidth]{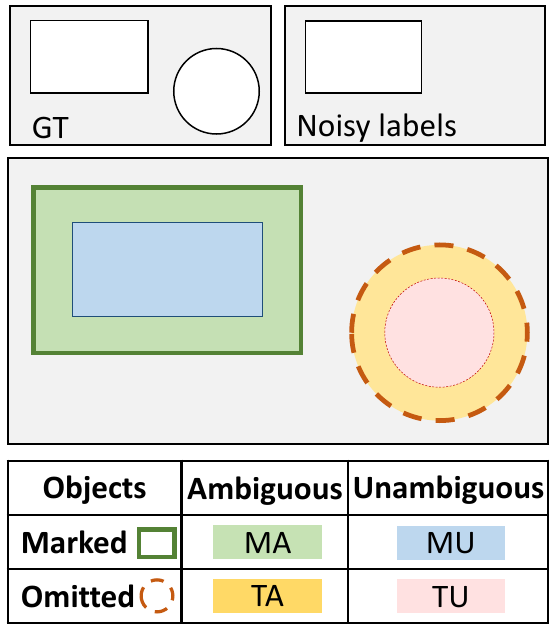} \vspace{0.1mm} &
        \includegraphics[width=1.1\linewidth]{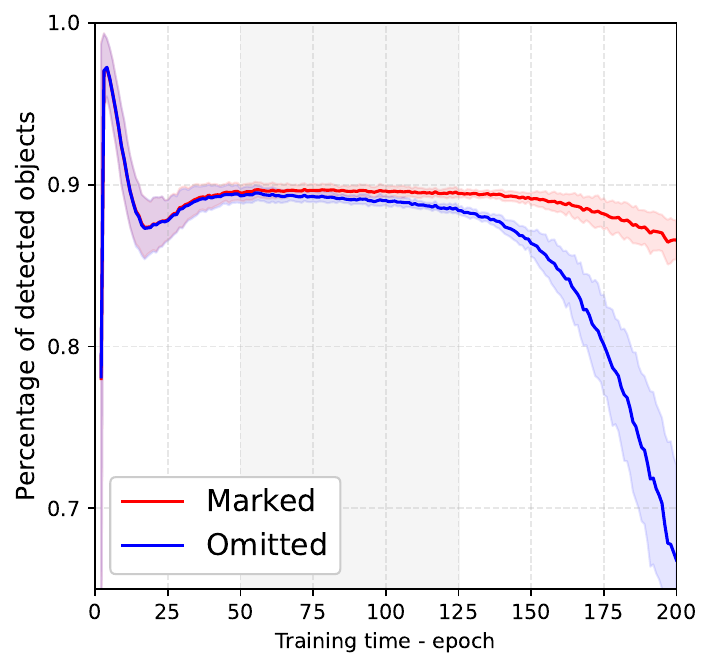} &
        \includegraphics[width=1.1\linewidth]{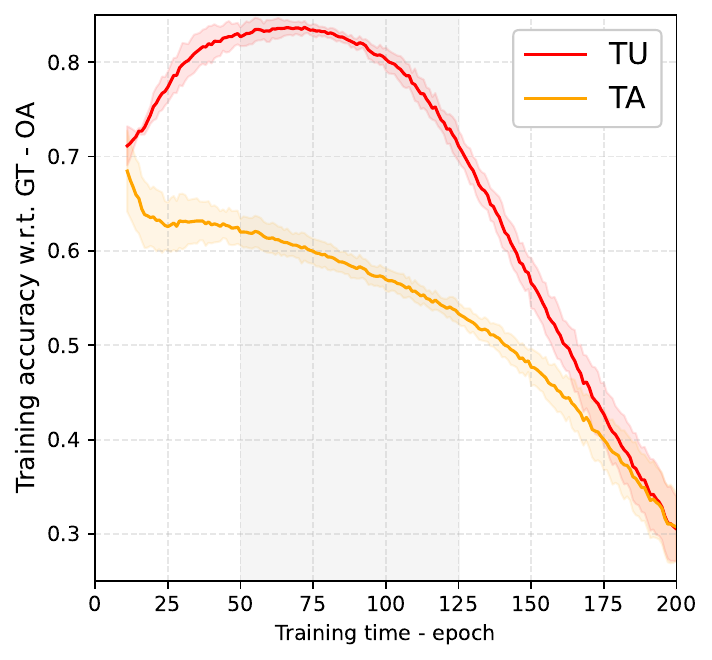} & 
        \includegraphics[width=1.1\linewidth]{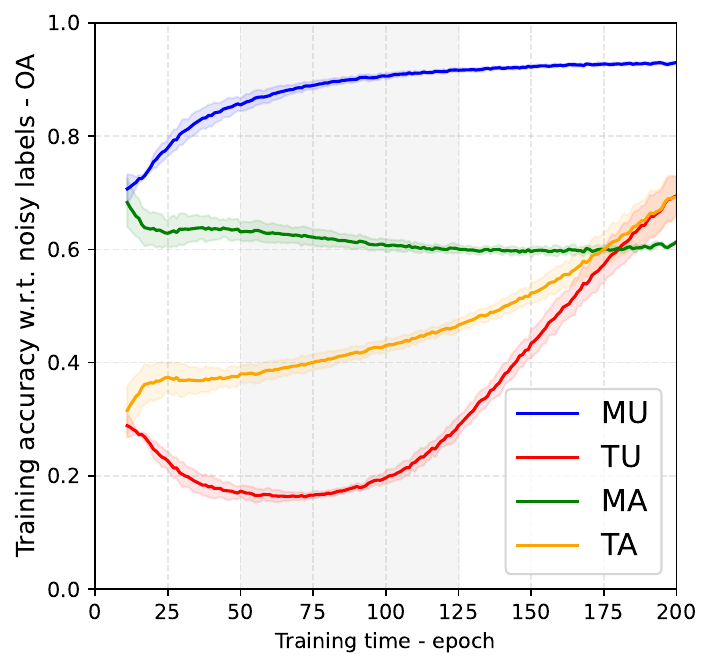} \\
        [\abovecaptionskip]
    \end{tabular}
    \begin{tabular}{m{0.22\linewidth}<{\centering}m{0.25\linewidth}<{\centering}m{0.19\linewidth}<{\centering}m{0.22\linewidth}<{\centering}}
        \small (a) Grouping strategy on GT &
        \small (b) Detection rates of objects &
        \small (c) OAs wrt\ GT &
        \small (d) OAs wrt\ noisy labels \\
    \end{tabular}
    \caption{{\textit{Numerical exploration of memorization effects on noisy labels for segmentation tasks (colors online)}: We statistically analyze the model training of the (teacher) model on the Massachusetts dataset at a drop rate of 0.5. %\newline
    From the object-wise perspective, we divide all the objects in ground-truth (GT) masks into Marked (solid \textcolor{ForestGreen}{green} rectangle) 
    and Omitted (dashed \textcolor{orange}{orange} circle), 
    and report their (b) detection rates. An object is rated \textit{detected} when it is at least partially predicted.
    From the pixel-wise perspective, we split all the object pixels into four groups as shown in (a), %: MA (\textcolor{ForestGreen}{green}), MU (\textcolor{blue}{blue}), TA (\textcolor{orange}{orange}), TU (\textcolor{red}{red}), 
    and calculate their overall accuracies (OAs) wrt (c) GT labels and (d) noisy labels. 
    %Note: the OAs of MA and MU are the same no matter what reference data is used. Therefore, we only present them in (d). 
    Gray background shadows highlight the transition phase.}} %identified leading into the memorization effects discussed.}}
    \label{fig:meth:memoAcc}
\end{figure*}

% seg maps - to show training dynamics
\begin{figure*}
    \centering
    \begin{tabular}{p{2.5cm}<{\centering}p{2.5cm}<{\centering}p{2.5cm}<{\centering}p{2.5cm}<{\centering}p{2.5cm}<{\centering}p{2.5cm}<{\centering}}
    \includegraphics[width=1.05\linewidth]{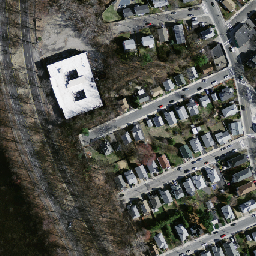} &
    \includegraphics[width=1.05\linewidth]{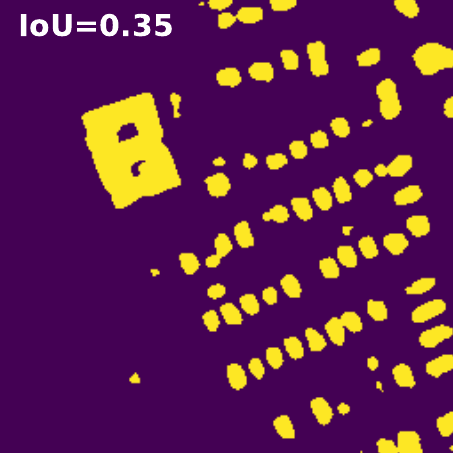} &
    \includegraphics[width=1.05\linewidth]{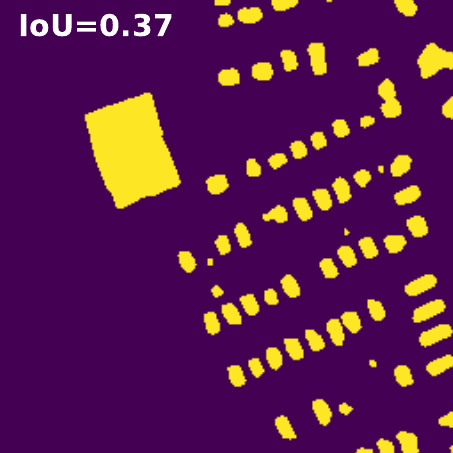} &
    \includegraphics[width=1.05\linewidth]{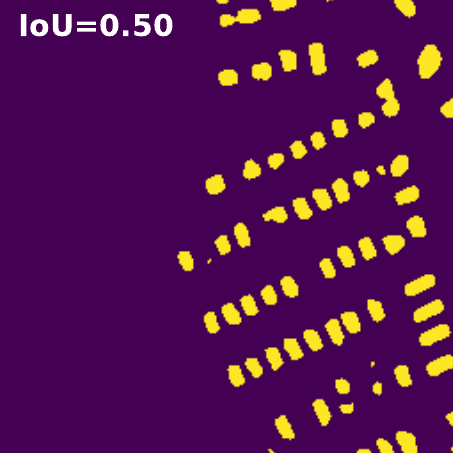} &
    \includegraphics[width=1.05\linewidth]{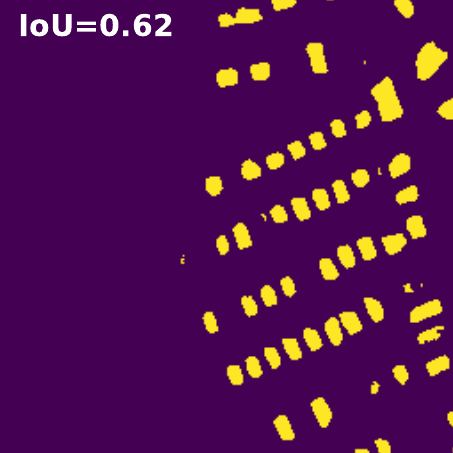} &
    \includegraphics[width=1.05\linewidth]{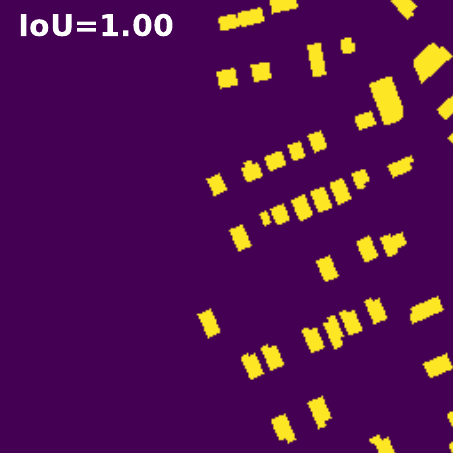} \\
    (a) Optical image & (b) 30th epoch & (c) 75th epoch & (d) 150th epoch & (e) 300th epoch & (f) Noisy labels
    \end{tabular}
    \caption{{\textit{An example demonstrating the three-stage training}: (b), (c), (d)-(e) show the predictions from the early-learning, transition, and memorization stages, respectively. We list the Intersection-over-Union (IoU) wrt (f) noisy labels at the upper left corner.}}
    % In the first early learning stage, e.g., at the (b) 30th epoch, model training is mainly about structure information and gradually refining details. In the second transition stage, e.g., at the (c) 75th epoch, models can extract most of the instances in the scene. In the last memorization stage, e.g., at the (d) 150th and (e) 300th epochs, models overfit to label noise at a rapid pace due to dominant learning of omitted-unambiguous samples (TU). {We list the Intersection-over-Union (IoU) wrt (f) noisy labels in the upper left corner in (b)-(f).}} }
    \label{fig:meth:segmap}
\end{figure*}

We first quantify the memorization effects from the object-wise perspective. We classify all objects\slash building footprints in the training set as either
\begin{itemize}
    \item \textbf{Marked} (M): identified as per the incomplete set, true positive (rectangle in \cref{fig:meth:memoAcc} (a)), or 
    \item \textbf{Omitted} (T): set of labels to complete the set, false negatives from the perspective of the incomplete set (circles in \cref{fig:meth:memoAcc} (a)).\end{itemize}
Note: As per the definition of \textit{incomplete labeling} false positives are not considered. \cref{fig:meth:memoAcc} (b) presents the detection rates. As the plot indicates, the transformation from early learning to memorization is smooth. After an initial unstable, oscillating warm-up phase, the model performance plateaus (transition stage),
%before gradually failing to detect omitted objects (transition stage) 
before the detection rate starts to sharply drop (memorization stage). 

At the pixel level we visualize the memorization phenomenon inspired by \cite{liu2022adele}, additionally considering the spatial correlation of pixels. Our grouping strategy is illustrated in \cref{fig:meth:memoAcc} (a). The pixels $P$ of a single object $O$ ($P\in O$) are broken down into two categories:
\begin{itemize}
    \item \textit{ambiguous} (A),  $P$ sufficiently close to the boundary $\partial O$ of $O$, $d(P,\partial O)<D$ for some maximum distance $D$ and distance function $d$ such as the Hausdorff metric \cite{dubuisson1994modified}
    \item \textit{unambiguous} (U), otherwise.
\end{itemize}
Given the previous object-level definitions of \textit{marked} and \textit{omitted}, all object pixels are categorized into one of the four groups: \textbf{marked-ambiguous (MA)}, \textbf{marked-unambiguous (MU)}, \textbf{omitted-ambiguous (TA)}, and \textbf{omitted-unambiguous (TU)}. \Cref{fig:meth:memoAcc} (c) and (d) present the evolution of the overall accuracy (OA) over the course of model training. 
{Note that the OAs of MA and MU are the same no matter what reference data is used. Therefore, they are only presented in \cref{fig:meth:memoAcc} (d).}
Due to memorization effects, we observe a notable bias as training progresses, i.e., while the OAs wrt ground-truth (GT) labels of TU and TA decrease in \cref{fig:meth:memoAcc} (c), their counterpart wrt noisy labels increase in \cref{fig:meth:memoAcc} (d). Yet, the memorization of TU and TA is out of sync. Noise memorization for TA takes place immediately upon the start of model training. In contrast, TU pixels stay unaffected to about epoch 75, as shown in \cref{fig:meth:memoAcc} (c), before overfitting reduces the OA wrt the GT. 
{As illustrated in \cref{fig:meth:segmap}, model training is mainly about structure information in the first early learning stage (see \cref{fig:meth:segmap} (b)). As a result, the model can extract most of the instances in the scene at the second transition stage (see \cref{fig:meth:segmap} (c)). However, in the memorization stage, the model overfits to label noise at a rapid pace, and learns to drop GT building footprints missing in the incomplete labels due to dominant learning of TU samples (see Figs. \ref{fig:meth:segmap} (d) and (e)).
%Over the course of training, as illustrated in \cref{fig:meth:segmap}, the model learns to drop GT building footprints missing in the incomplete labels.
}
Consequently, the OA wrt GT drops while the OA wrt noisy labels ramps up. On the other hand, it seems that the boundary regions of building footprints are more vulnerable to overfitting. Indeed, properly defining the exact outline of a building from aerial imagery is a challenging task, even for humans. For example, do annotations follow subtle details of the building's facade, or is the whole footprint approximated by a simplified rectangle? Are open courtyards (with green areas) considered part of the building footprint? We observe that overfitting reflects most notably in TA pixels, resulting in coarse approximation of building footprint outlines before memorization affects the inner core of objects semantically segmented (see Figs. \ref{fig:meth:segmap} (b) and (c)).

To summarize, as illustrated in \cref{fig:meth:memo3stages}, {on both datasets,} in the first early-learning stage, dominant learning of MU leads to rapid increase of training accuracies wrt both GT and noisy labels. Then in the second transition stage, the learning of MU is close to an end--it is mainly TA samples that are being memorized. The training arrives at a plateau where both kinds of training accuracies keep relatively stable. In the final memorization stage, the training accuracies wrt noisy labels again start to increase rapidly while training accuracies wrt GT drop severely, largely due to the ultimate memorization of TU samples. {In practice, the number of epochs representing different stages would vary when the degree of incompleteness, the image resolution, as well as the tasks (e.g., road detection, urban green space detection, etc.) change as indicated in \cref{fig:meth:memo3stages} (a) and (b).}

In the next section, based on the analysis of the three-stage training, we introduce the ACT and O2C modules to solve the ``when'' and ``how'' questions in our label correction pipeline.

\subsection{Adaptive Correction Trigger Module} \label{sec:meth:act}

\begin{figure}
    \centering
        \includegraphics[width=0.90\linewidth]{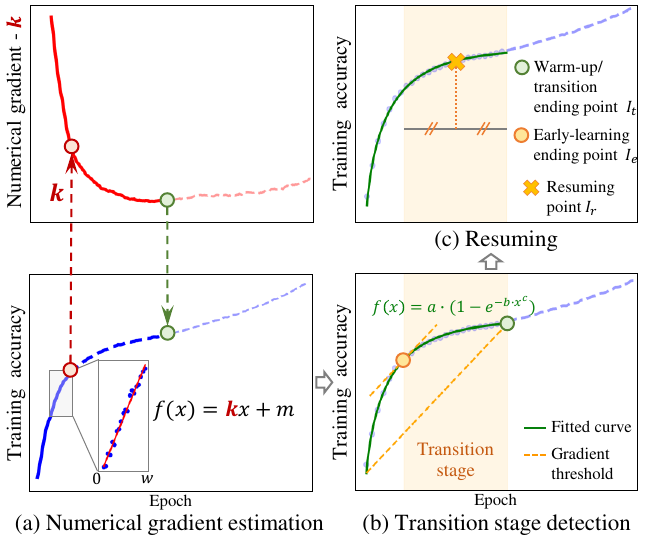} 
    \caption{{\textit{Adaptive Correction Trigger (ACT) module}: a three-stage strategy to identify ``when'' to start label correction (\textcolor{Orange}{orange $\times$}), where the non-negative number $w$ in (a) determines the window size of epochs to numerically estimate $k$, the \textcolor{Blue}{blue} faded dashed line (\textcolor{Blue}{$--$}) indicates trends in the training accuracy obtained without the application of our ACT module.
    }
    }
    %After an initial warm-up and slow-down in training accuracy, a saddle point in training accuracy toward re-acceleration is a hallmark of data overfitting\slash\textit{memorization} taking effect (\textcolor{Green}{green $\circ$}). As per our experiments, the minimum in training accuracy rate, numerical gradient $k$ (top left plot), marks the epoch $I_t$ ending a transition stage (\textcolor{amber}{yellow shaded area}) of learning structural information for semantic segmentation, as shown in \cref{fig:meth:segmap}. We define the beginning of the transition stage (\textcolor{Orange}{orange $\circ$}) as the epoch $I_e$, where the rate of training accuracy increase matches the overall rate of training accuracy increase from epoch zero to $I_t$. The non-negative number $w$ in the inset of the lower left panel determines the window size of epochs to numerically estimate $k$. Note: the \textcolor{Blue}{blue} faded dashed line (\textcolor{Blue}{$--$}) indicates trends in the training accuracy obtained without the application of our ACT module.}} %never fitted to in our methodology.}}
    \label{fig:meth:act}
\end{figure}

As stated above, the model reaches the highest potential in the transition stage when directly trained with noisy labels. A natural idea to solve the ``when'' question is to initiate the correction procedure in this stage, using the most reliable predictions. As shown in \cref{fig:meth:memo3stages}, the training accuracy (wrt noisy labels) increases much faster both before and after the transition stage, which provides us the potential to monitor the growth rate of the training accuracy curve for detection.
{
\Cref{fig:meth:act} showcases the detection procedure of the ACT module following this idea. After an initial warm-up and slow-down in training accuracy, a saddle point in training accuracy toward re-acceleration is a hallmark of data overfitting\slash\textit{memorization} taking effect (\textcolor{Green}{green $\circ$}), which marks the ending of the transition stage (\textcolor{amber}{yellow shaded area}) of learning structural information for semantic segmentation. We define the beginning of the transition stage (\textcolor{Orange}{orange $\circ$}) as the epoch $I_e$, where the rate of training accuracy increase matches the overall rate of training accuracy increase from epoch zero to $I_t$. The resuming point from which the label correction is triggered is finally taken as the middle point of the transition stage.
}

To numerically determine the gradient of the training accuracy over epochs, we apply local linear fitting over $w$ epochs based on the outputs of the teacher network. This approach reduces random fluctuations in the estimation of the training accuracy's derivative of the student model. Given the sliding window size $w$, the numerical gradient at the $i$-th epoch is estimated by fitting the $(i-w)$-th to the $i$-th training accuracy data points $\{(x,y)\}_w=\{(1,f_{i-w}),\dots (w,f_i)\}$ to a linear function
\begin{equation} \label{eq:act:linear}
    f(x)=k_ix+m_i=y\quad,
\end{equation}
where the slope parameter $k_i$ represents the growth rate of training accuracy at each epoch $i$. 

The second step is to determine the transition stage. We expect $k$ to continuously decrease in the first two stages, while it ramps up again when entering the last memorization stage. So we terminate the warm-up phase at the end of the transition stage when $k$ starts to increase. Let $z$ be the \textit{look-ahead buffer zone size} to determine whether $k_i$ has hit its lowest numerical value. Then the ending point of warm-up/transition stage $I_t$ is detected as
\begin{equation} \label{eq:act:tranend}
    I_t = j \quad \text{when} \quad k_j=\min(k_j,\cdots,k_{j+z})\quad.
\end{equation}
To improve the robustness of detection, we perform the analysis, \cref{eq:act:tranend}, for a set of sliding window sizes $W=\{w\}$, i.e., we obtain the set $I=\{I_t^{(w)}\}_W$ over which we average accordingly:
\begin{equation} \label{eq:act:endp} \footnotesize
\langle I_t\rangle = \frac{1}{\vert W\vert}\left\lfloor\sum_{w\in W}I^{(w)}_t\right\rfloor\quad,
\end{equation}
with $\vert W\vert$ the number of window sizes picked. Details on the choice of $w$ values are documented in \cref{sec:exp:set}. The buffer parameter $z$ is simply set to the mean $z=\langle w\rangle=\lfloor w\rfloor$.

Thereafter, to detect the ending point of the early-learning stage or the starting point of the transition stage $I_e$, we need a threshold to tell when the curve becomes sufficiently flat. A manually set threshold depends heavily on the quality of noisy labels, which is hard to fix in most scenarios. Alternatively, we propose that the slope between the first and $I_t$-th epochs serve as the adaptive threshold (the orange lines in \cref{fig:meth:act}), that is, we compute the slope of the arc as
\begin{equation} \label{eq:act:thr} \footnotesize
    \sigma = \frac{f_{I_t}-f_1}{I_t}\quad. % (f_{I_t}-f_1)/I_t % \frac{f_{I_t}-f_1}{I_t}
\end{equation}
At this step we turn to a different evaluation of the gradients compared to \cref{eq:act:linear} above, now fitting a multi-parameter curve, because
\begin{itemize}
    \item we have sufficient data points from the warm-up phase to globally fit the exponentially saturating function \cref{eq:act:exp}, and
    \item its analytic gradients \cref{eq:act:expgrad} are monotonically decreasing.
\end{itemize}
In this sense, we fit $I_t$ training accuracy points $\{(x,y)\}_{I_t}=\{(1,f_1),\dots (I_t,f_{I_t})\}$  to
\begin{equation} \label{eq:act:exp}
    f(x)=a\left(1-\exp{(-bx^c)}\right)=y\quad,
\end{equation}
with $a$, $b$, and $c$ fitting parameters. Accordingly, we obtain 
{
\begin{equation} \label{eq:act:expgrad}
    f'(x)=abc\ x^{c-1}\exp{(-bx^c)}\quad.
\end{equation}
}
Specifically, $0<a<1$ corresponds to the magnitude of \cref{eq:act:exp}. We constrain $b>0$ and $0<c<1$ to restrict \cref{eq:act:expgrad} to a monotonically decreasing function. Correspondingly, employing the threshold, \cref{eq:act:thr}, we can explicitly count to the starting point of the transition phase as follows:
{
\begin{equation} \label{eq:act:elend} 
    % I_e = \sum_{i=1}^{I_t}\left[f'(i)>\sigma\right]\quad,
    I_e = \sum_{i=1}^{I_t}\text{sgn} \big(f'(i)-\sigma\big)\quad,
\end{equation}
where $\text{sgn}(\cdot)$ is the sign function with $\text{sgn}(x)=1$ if $x>0$, otherwise $0$, and $\sigma$ is the adaptive threshold defined in \cref{eq:act:thr}.
}
%where the term in square brackets evaluates to numerical value $1$ if the statement inside is true, and $0$, otherwise.

After determining the values of $I_e$ and $I_t$, we finally resume the models at the middle of the transition stage that trigger O2C from the $I_r$-th epoch with
\begin{equation} \label{eq:act:ir} \footnotesize
    I_r=\left\lfloor\frac{I_e+I_t}{2}\right\rfloor\quad,
\end{equation}
which is expected to be close to the best-performed model in the warm-up phase. However, saving every checkpoint file consumes a lot of storage space. A trick here is to save a selection of checkpoints, e.g., every 5 checkpoints, and resume from the one closest to $I_r$. 

\subsection{Online Object-wise Label Correction} \label{sec:meth:o2c}

\begin{figure}
    \centering
        \includegraphics[width=0.9\linewidth]{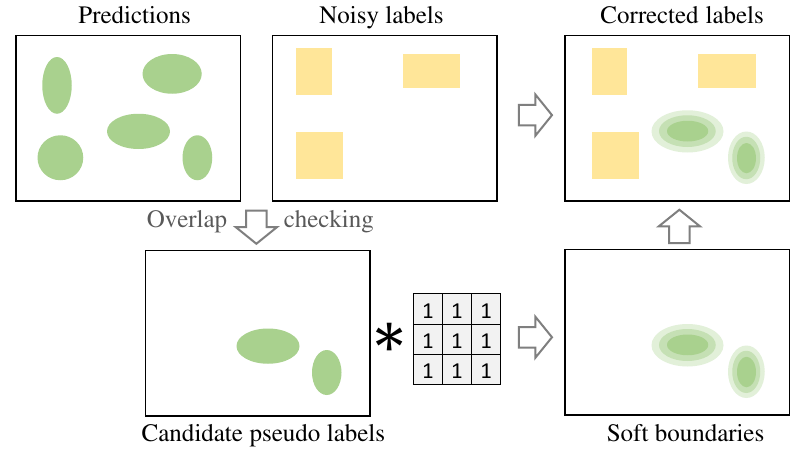} 
    \caption{{\textit{Online Object-wise label Correction (O2C) module}: A spatial constraint is used to solve ``how" to correct labels, where candidate pseudo labels are selected by filtering predictions (solid \textcolor{Green}{green} ellipse) against noisy labels (\textcolor{Yellow}{yellow} rectangles), and
    $\ast$ is the convolution operator to introduce uncertainty around boundary areas.}}
    %Predicted labels (solid \textcolor{Green}{green} ellipse) are filtered against the noisy labels (\textcolor{Yellow}{yellow} rectangles). Predictions not matching the noisy labels' spatial extent are added as \textit{corrected} (top right panel) after convolution of a rasterized binary segmentation map (\texttt{0}\dots no label, \texttt{1}\dots label) with, e.g., a 3x3 square kernel of values \texttt{1} (lower left panel). Effectively, this operation marks boundary pixels of a labeled object less certain compared to the bulk of the object, as shown by blurry \textcolor{Green}{green} ellipses in right column.}}
    \label{fig:meth:o2c}
\end{figure}

As discussed in \cref{sec:meth:memo} and shown in \cref{fig:meth:memoAcc}, the memorization of noisy labels mainly takes place on TA samples around boundaries during the transition stage with a high object detection rate of model. Based on this observation, we design an Online Object-wise label Correction (O2C) module as a substitute for the commonly used pixel-wise correction strategies. 

\Cref{fig:meth:o2c} presents the workflow of O2C. One major improvement is the selection of pseudo label candidates in an object-wise fashion by checking the overlap between predictions and given noisy labels. We reserve the marked objects, and do label correction only for those that are omitted. In addition, considering the memorization effects on TA samples, we apply a smooth filter to generate soft boundaries for candidate pseudo objects. For simplicity, we use an all-one filter in \cref{fig:meth:o2c}, which can also be replaced by other kinds of smooth filters, such as a Gaussian filter.

Additionally, ``online'' in the O2C name includes one-off label correction at each iteration without saving historical correction results. This is another major difference from commonly used pixel-wise correction strategies, which correct labels incrementally.

\begin{algorithm}
\small
\caption{{AIO2 Framework}}\label{alg:aio2}
\hspace*{\algorithmicindent} 
\textbf{Input}: training set with noisy labels $D_T({\mathbf x},y)$, look-ahead buffer zone size $z$ and a set of sliding window sizes $W={w}$ for sample correction detection \\
\hspace*{\algorithmicindent} 
\textbf{Output}: predicted segmentation masks
\begin{algorithmic}[1]
\STATE {\bf Initialization}: randomly initialized student model $\theta^{(s)}_{0}$ with teacher model $\theta^{(t)}_{0}$=$\theta^{(s)}_{0}$ and $\theta^{(t)}_{0}$.requires$\_$grad=False, empty accuracy list of teacher model $A$, empty numerical gradient lists $K^{(w)}$, indicator for sample correction $S$=False
\FOR{$i$=1 to Epoch}
    \FOR{$j$=1 to Batch}
        \STATE $\textsc{//}\texttt{O2C\;for\;label\;correction}$
        \IF{$S$}
            \STATE $\Tilde{y}=\text{O2C}(y,\theta^{(t)}_{i-1}({\bf x}))$, cf. \cref{sec:meth:o2c}
        \ELSE
            \STATE $\Tilde{y}=y$
        \ENDIF {$\;S$}
        \STATE $\textsc{//}\texttt{model\;updating}$
        \STATE update student model $\theta^{(s)}_{i(j)}$ with $({\bf x}, \Tilde{y})$, cf. Eqs. (\ref{eq:teacher:sup}), (\ref{eq:loss})
        \STATE update teacher model $\theta^{(t)}_{i(j)}$ with $\theta^{(s)}_{i(j)}$, cf. \cref{eq:teacher:tup}
    \ENDFOR {$\;i$}
    \STATE $\textsc{//}\texttt{ACT\;for\;label\;correction\;detection}$
    \IF{not $S$}
        \STATE update $A=A+[\text{IoU}(D_T,\theta_{i}^{(s)})]$
        \STATE update $K^{(w)}=K^{(w)}+[\text{FIT}(w,i,A)]$ cf. \cref{eq:act:linear} 
        \IF{$k_{i-z}^{(w)}$ for $w$ in $W$ all meet \cref{eq:act:tranend}}
            \STATE get the resuming point $I_r$, cf. Eqs. (\ref{eq:act:endp})-(\ref{eq:act:ir})
            \STATE $\textsc{//}\texttt{resume\;from\;the\;$\texttt{I}_r$-th\;epoch}$\\
            \hspace*{\algorithmicindent}$\texttt{and\;trigger\;label\;correction}$
            \STATE $i\leftarrow i$-$z$, $\theta^{(s)}_{i} \leftarrow \theta^{(s)}_{i-z}$, $\theta^{(t)}_{i} \leftarrow \theta^{(t)}_{i-z}$, $C \leftarrow$ True
        \ENDIF {$\;k_{i-z}^{(w)}$}
    \ENDIF {$\;$not $S$}
\ENDFOR {$\;i$}
\end{algorithmic}
\end{algorithm}

\subsection{{AIO2 Framework}}
{In summary, the proposed AIO2 framework employs a two-stage pipeline to train segmentation networks with incomplete noisy labels. In the initial warm-up stage, models are trained using the given noisy labels. Subsequently, in the second stage, O2C is applied for label correction before optimization. The transition between the two stages is automatically managed by ACT. It is crucial to notice that the teacher model plays a predominant role in both the ACT and O2C modules, while solely the student model is directly trained with training data and gradient descent. We summarize the implementation details in \cref{alg:aio2} for a more comprehensive understanding of AIO2. Specifically, within each epoch, the student model undergoes the optimization with the training set, using a combined loss of the distribution-based cross-entropy loss and the region-based dice loss \cite{jadon2020asurvey}, as follows,
\begin{equation} \label{eq:loss} 
    L = L_{ce}+L_{dice}\quad, 
\end{equation}
\begin{equation} \small
    L_{ce} = -\sum_{i=1}^{N}\sum_{j=1}^{C}y_{i,j}\log(p_{i,j})\quad,
\end{equation}
\begin{equation} \small
    L_{dice} = 1-\frac{2\sum_{j=1}^{C}\sum_{i=1}^{N}{y_{i,j}p_{i,j}}}{\sum_{j=1}^{C}\sum_{i=1}^{N}(y_{i,j}+p_{i,j})}\quad,
\end{equation}
where $y_{i,j}$ represents the one-hot label for the $i$th sample at class $j$, $p_{i,j}$ denotes the corresponding prediction from the softmax layer, $N$ and $C$ are the numbers of samples and classes. After that, the teacher model is updated by EMA with the new weights of the student model, as opposed to traditional gradient descent. 
}

% data example - Boston
\begin{figure}
    \centering
    \footnotesize
    \begin{tabular}{p{2.3cm}<{\centering}p{2.3cm}<{\centering}p{2.3cm}<{\centering}}
        \includegraphics[width=1.05\linewidth]{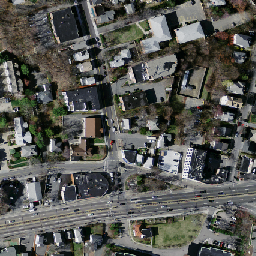} &  
        \includegraphics[width=1.05\linewidth]{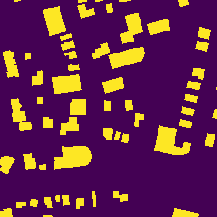} &
        \includegraphics[width=1.05\linewidth]{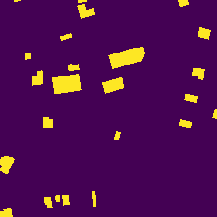} \\
        \includegraphics[width=1.05\linewidth]{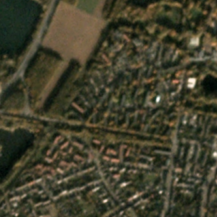} &  
        \includegraphics[width=1.05\linewidth]{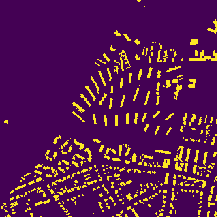} &
        \includegraphics[width=1.05\linewidth]{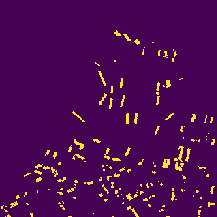} \\
        (a) Optical image & (b) GT & (c) Noisy labels
    \end{tabular}
    \caption{Example of data triples for two datasets. (b) Accurate ground-truth (GT) labels were used to generate (c) noisy labels with the designed label noise injection strategy. The first and second rows correspond to the Massachusetts (1m) and Germany (3m) datasets, respectively. Buildings are highlighted in yellow in (b) and (c).}
    \label{fig:data}
\end{figure}

\section{Experiments} \label{sec:exp}

In this section, we evaluate the effectiveness of our proposed AIO2 method on two building footprint extraction datasets with different spatial resolutions. We first give an overview of the two datasets and label noise injection strategy along with other settings in \cref{sec:exp:set}, followed by detailed experimental results, including ablation studies and parameter sensitivity analysis, and related discussions.  

\subsection{Datasets and Settings} \label{sec:exp:set}

\subsubsection{Datasets} \label{sec:exp:set:data}

The \textbf{Massachusetts Dataset}\footnote{Downloaded from \url{https://www.cs.toronto.edu/~vmnih/data/}.} \cite{Mnih2013Thesis} is composed of 151 RGB aerial images collected over the City of Boston with a size of 1500 $\times$ 1500 pixels and a spatial resolution of 1m. The corresponding building masks are also provided for each image. The whole dataset was randomly split into three subsets comprising a training set of 137 images, a test set of 10 images, and a validation set of 4 images. We keep the original split, and further crop each image into a series of 256 $\times$ 256 small patches. After some images without labels are removed, the final datasets comprise 3065 patches for training, 250 for test, and 100 for validation.

The \textbf{Germany Dataset} \cite{Skuppin2022igarss} consists of 2052 image-label pairs with a size of 320 $\times$ 320 pixels generated across ten Germany cities including Bielefeld, Bochum, Bonn, Cologne, Dortmund, Duesseldorf, Duisburg, Essen, Muenster, and Wuppertal. The image data were from Planet basemap images with a lower spatial resolution of 3m and 3 bands (RGB). The building masks were rasterized from vector cadastral data\footnote{ Accessed via GEOportal.NRW (\url{https://www.geoportal}) on Aug. 5, 2021.} to 3m GSD to pair with image data. In our experiments, we randomly select 200 patches for test and 50 for validation. 

\subsubsection{Label noise injection strategy} \label{sec:exp:set:noise_inject}
We inject incomplete label noise by randomly discarding a certain proportion of instances from ground-truth (GT) building masks, as illustrated in \cref{fig:data}. Let $\alpha$ denote the discarding percentage. To simulate the inconsistency in sample quality among different local areas, we perform uniform sampling from the range centered on the given discarding percentage for the whole dataset $\alpha_0$ to obtain the actual $\alpha$ for each patch. The range is defined as $[\alpha-r,\alpha+r]$ with $r=\min(1-\alpha,\alpha)$. For example, we sample $\alpha$ from $[0,0.6]$, $[0,1]$, $[0.4,1]$ given $\alpha_0=0.3, 0.5, 0.7$, respectively. We implement 3 replays under each $\alpha_0$ in our experiments to show the statistical significance of results. The quality assessment of our synthetic noisy labels is presented in \cref{tab:exp:quality}. As can be observed, the overall omission rates are close to $\alpha_0$ though $\alpha$ differs on each patch.

% quality assessment of two datasets
% \renewcommand{\arraystretch}{1.2}
\begin{table}[t]
    \begin{center}
    \begin{threeparttable}
    \caption{Quality assessment of synthetic noisy labels given different discarding percentages $\alpha_0$ on two datasets.}%\tnote{\dag}.}%, where the standard deviations were calculated from three replays with different random seeds. The three assessment metrics used here are Omission Rate (OR), denoting the actual instance discarding percentages of noisy labels versus GT, Intersection over Union (IoU) of building class, and Overall Accuracy (OA).}
    \label{tab:exp:quality}
    \small
    \begin{tabular}{c|c|c|c|c}
        \hline\hline
        % Dataset & \multicolumn{3}{c?}{Massachusetts} & \multicolumn{3}{c}{Germany} \\
        % \hline
        \multicolumn{2}{c|}{$\alpha_0$} & 0.3 & 0.5 & 0.7 \\
        \hline\hline
        \multirow{3}{*}{Massachusetts} & OR & 29.48$\pm$0.29 & 49.40$\pm$0.86 & 69.46$\pm$0.29  \\
        & IoU & 71.35$\pm$0.46 & 51.58$\pm$0.85 & 31.71$\pm$0.37 \\
        & OA & 96.17$\pm$0.06 & 93.52$\pm$0.11 & 90.86$\pm$0.05 \\
        \hline\hline
        \multirow{3}{*}{Germany} 
        & OR & 29.75$\pm$0.58 & 49.31$\pm$0.60 & 69.58$\pm$0.36  \\
        & IoU & 70.35$\pm$0.43 & 50.69$\pm$0.56 & 30.29$\pm$0.36 \\
        & OA & 96.78$\pm$0.02 & 94.62$\pm$0.07 & 92.40$\pm$0.03 \\
        \hline\hline
    \end{tabular}
    
    \scriptsize{
    % \begin{tablenotes}
    %\item[\dag] 
    Note: The standard deviations were calculated from three replays with different random seeds. The three assessment metrics used here are Omission Rate (OR) denoting the actual instance discarding percentages of noisy labels versus GT, Intersection over Union (IoU) of building class, and Overall Accuracy (OA).}
    % \end{tablenotes}
    \end{threeparttable}
    \end{center}
\end{table}

\subsubsection{Models and compared methods} \label{sec:exp:set:cmp_mthd}
We utilize U-Nets \cite{ronneberger2015unet} as our building extraction models with vanilla U-Net ecoder and EfficientNet B5 \cite{tan2019efficientnet} as backbones for the Massachusetts and Germany datasets, respectively. 
We have seven methods to compare in total, including two baselines (U-Nets directly trained with GT and noisy labels) and five other methods, two based on pixel-wise label correction and three using regularization techniques.
{The baseline results of training with GT can be taken as the potential upper limit of these sample correction methods.}

For the pixel-wise label correction, a common approach is to apply a fixed threshold $K$ on confidence values (softmax outputs) to select correction candidates \cite{liu2022adele, cao2023building}. We refer to this approach as \textbf{pixel-wise}, and choose $K=0.6$ as the threshold value after a parameter tuning. In addition, we compare our proposed approach with an \textbf{adaptive pixel-wise} version, which automatically sets the thresholds for each patch as the minimum between $K$ and its averaged confidence value on this patch \cite{dong2022landcover}. Furthermore, we also upgrade it by incorporating class-wise thresholds for better performance. To ensure the effectiveness of pixel-wise correction methods, we combine them with our designed ACT module, and take teacher model outputs as corrected pseudo labels.

In addition, we employ three regularization techniques: \textbf{consistency constraint} \cite{tarvainen2017meanteacher, wang2020consist}, which enforces consistency between teacher and student model outputs; \textbf{bootstrapping} \cite{Reed2015boot, Henry2021boot}, which combines original noisy labels and model predictions as soft reference data for loss calculation; and \textbf{noisy label regularization} \cite{cao2022green, dong2022landcover}, which adds a weighted loss with respect to the original noisy labels in the adaptive pixel-wise training scheme. The consistency constraint is formulated in the form of mean squared error (MSE) with an adaptive weight gradually ramping up to 0.7 in the first 80 epochs. Bootstrapping combines soft pseudo labels (softmax outputs) $\hat{p}$ and original hard noisy labels $y$ by $y'=\beta\cdot y+(1-\beta)\cdot \hat{p}$, with $\beta$ exponentially decreasing from 1 to 0.3 in the first 80 epochs. The weight for noisy label regularization is set as 0.25 via parameter searching.

For the proposed AIO2 method, we set the sliding window size $w$ group in ACT as $[10,20,30,40]$ for early learning detection. Taking into account the spatial resolution of the two datasets, we select 5 and 3 as the filter size to generate soft boundaries of pseudo labels in O2C on the Massachusetts (1m) and Germany (3m) datasets, respectively. Adam serves as the optimizer in all methods with a learning rate of 1e-3 and 5e-3 for the Massachusetts and Germany datasets, respectively. Since we assume no clean labels to be available during training in the noisy label settings, we only use the validation set to select models when trained with GT. The final results reported for the other 7 LNL methods derive from 3 replays after 325 and 300 epochs on the Massachusetts and Germany datasets, respectively. Our evaluation measures include Intersection over Union (IoU), precision (the positive fraction among predictions), recall (the positive fraction among GT labels/those supposed to be positive), and F1 score (the harmonic mean of the precision and recall) of the building class, along with Overall Accuracy (OA).

% Boston - comparison result figures
\begin{figure*}[htp]
    \centering
    \small
    \includegraphics[width=0.95\linewidth]{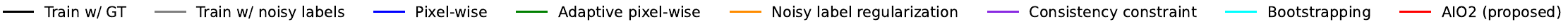}
    \begin{tabular}{p{5.5cm}<{\centering}p{5.5cm}<{\centering}p{5.4cm}<{\centering}}
        \includegraphics[width=1.\linewidth]{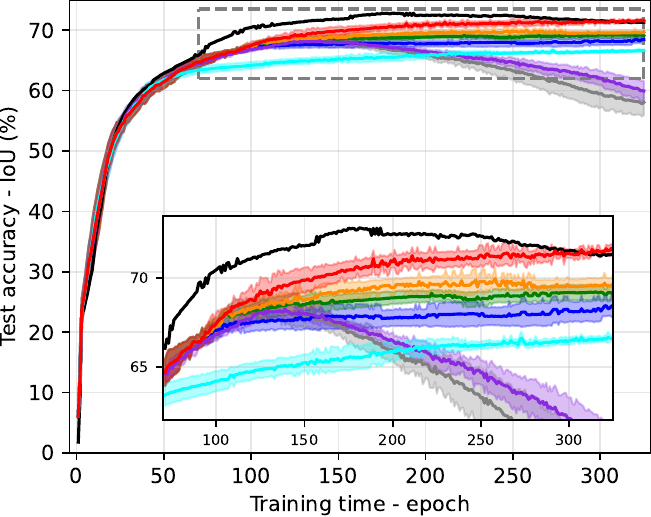} &  
        \includegraphics[width=1.\linewidth]{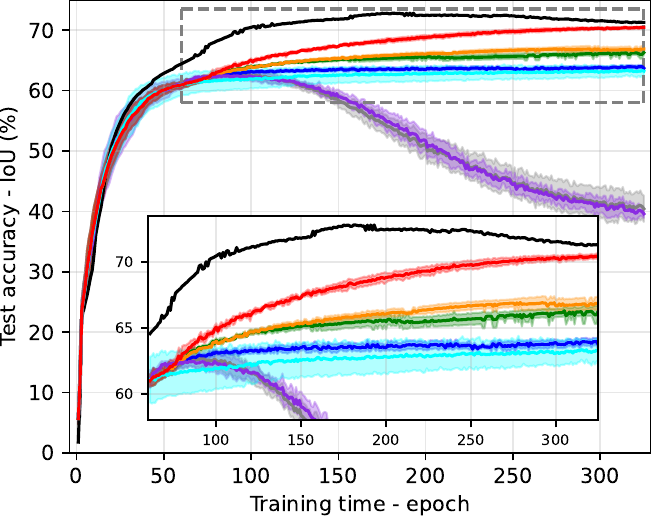} &
        \includegraphics[width=1.\linewidth]{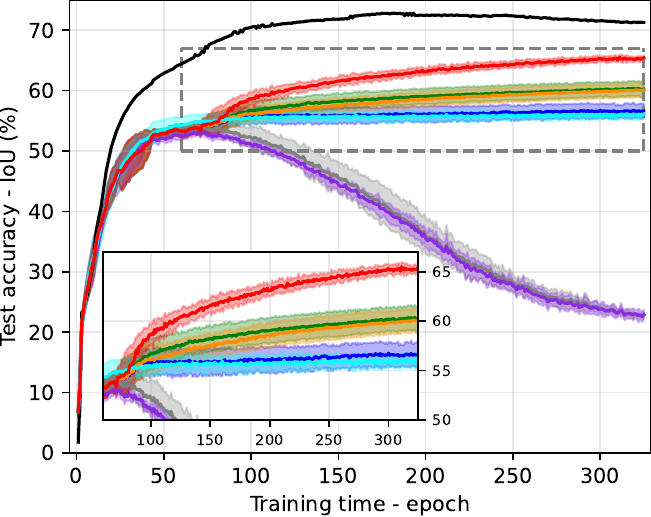} \\
        \hspace{0.7cm} (a) $\alpha_0$ = 0.3 & 
        \hspace{1cm} (b) $\alpha_0$ = 0.5 & 
        \hspace{1cm} (c) $\alpha_0$ = 0.7
    \end{tabular}
    \caption{Test accuracy (IoU) versus training time (epoch) obtained by considered methods trained with GT or incomplete noisy labels under different given discarding percentages ($\alpha_0$) on the Massachusetts dataset.}
    \label{fig:exp:hsota}
\end{figure*}

% Boston - comparison result IoU tables
\renewcommand{\arraystretch}{1.2}
\begin{table*}[htp]
    \begin{center}
    \scriptsize
    \caption{IoUs obtained by considered methods after 325 epochs on the Massachusetts dataset.}%, where the best and second best results among LNL methods are highlighted in bold and underlined, respectively.}
    \label{tab:exp:hsota-iou}
    \begin{tabular}{c|c?c|c|c?c|c|c}
    \hline\hline
    \multicolumn{2}{c?}{IoU (\%)} & \multicolumn{3}{c?}{Final} & \multicolumn{3}{c}{Maximum} \\
    \hline
    \multicolumn{2}{c?}{$\alpha_0$} & 0.3 & 0.5 & 0.7 & 0.3 & 0.5 & 0.7 \\
    \hline\hline
    \multirow{2}{*}{Baseline} & Train w/ GT & \multicolumn{3}{c?}{72.47} & \multicolumn{3}{c}{72.83} \\
    \cline{2-8}
    & Train w/ noisy labels & 58.06$\pm$2.13 & 40.43$\pm$2.17 & 22.84$\pm$0.48 & 68.51$\pm$0.50 & 62.82$\pm$0.60 & 54.22$\pm$0.82 \\
    \hline\hline
    \multirow{3}{*}{Regularization} 
    & Consistency constraint & 59.94$\pm$1.69 & 39.59$\pm$0.66 & 22.90$\pm$0.86 & 68.31$\pm$0.38 & 62.85$\pm$0.28 & 53.35$\pm$0.54 \\
    \cline{2-8}
    & Bootstrapping & 66.63$\pm$0.19 & 63.23$\pm$1.01 & 56.11$\pm$0.58 & 66.83$\pm$0.26 & 63.52$\pm$0.96 & 56.33$\pm$0.75 \\
    \cline{2-8}
    & Noisy label regularization & \underline{69.71$\pm$0.58} & \underline{66.88$\pm$0.51} & 60.09$\pm$1.14 & \underline{70.10$\pm$0.52} & \underline{67.10$\pm$0.44} & 60.15$\pm$1.19 \\ 
    \hline\hline
    \multirow{3}{*}{Correction} 
    & Pixel-wise & 68.45$\pm$0.60 & 63.71$\pm$0.42 & 56.63$\pm$1.18 & 68.46$\pm$0.60 & 64.00$\pm$0.28 & 56.85$\pm$1.22 \\
    \cline{2-8}
    & Adaptive pixel-wise & 69.06$\pm$0.31 & 66.29$\pm$0.35 & \underline{60.25$\pm$1.21} & 69.41$\pm$0.20 & 66.37$\pm$0.36 & \underline{60.42$\pm$1.24} \\
    \cline{2-8}
    & AIO2 (proposed) & \textbf{71.56$\pm$0.29} & \textbf{70.47$\pm$0.17}  &  \textbf{65.45$\pm$0.27} & \textbf{71.72$\pm$0.27} & \textbf{70.50$\pm$0.19} & \textbf{65.62$\pm$0.39} \\
    \hline\hline
    \end{tabular}
    \end{center}
    \scriptsize{\hspace{0.8cm} Note: the best and second best results among LNL methods are highlighted in bold and underlined, respectively.}
\end{table*}
\renewcommand{\arraystretch}{1}

% Boston - comparison result accuracy tables
\renewcommand{\arraystretch}{1.2}
\begin{table*}[htp]
    \begin{center}
    \scriptsize
    \caption{OAs, precisions, recalls, and F1 scores obtained by considered methods after 325 epochs on the Massachusetts dataset.}%, where the best and second best results among LNL methods are highlighted in bold and underlined, respectively.}
    \label{tab:exp:hsota-acc}
    \begin{tabular}{c|c?c|c|c?c|c|c?c|c|c?c|c|c}
    \hline\hline
    \multicolumn{2}{c?}{Accuracy (\%)} & \multicolumn{3}{c?}{OA} & \multicolumn{3}{c?}{Precision} & \multicolumn{3}{c?}{Recall} & \multicolumn{3}{c}{F1} \\
    \hline
    \multicolumn{2}{c?}{$\alpha_0$} & 0.3 & 0.5 & 0.7 & 0.3 & 0.5 & 0.7 & 0.3 & 0.5 & 0.7 & 0.3 & 0.5 & 0.7\\
    \hline\hline
    \multirow{2}{*}{Baseline} & Train w/ GT & \multicolumn{3}{c?}{94.86} & \multicolumn{3}{c?}{85.23} & \multicolumn{3}{c?}{83.19} & \multicolumn{3}{c}{84.20} \\
    \cline{2-14}
    & Train w/ noisy labels & 92.14 & 88.64 & 85.07 & \underline{86.60} & \underline{86.94} & \underline{87.10}  & 64.39 & 43.51 & 23.88 & 73.84 & 57.97 & 37.49 \\
    \hline\hline
    \multirow{3}{*}{Regularization} 
    & Consistency constraint & 92.40  & 88.58 & 85.09 & \textbf{87.50} & \textbf{87.78} & \textbf{88.56} & 66.24 & 42.33 & 23.85 & 75.37 & 57.11 & 37.57 \\
    \cline{2-14}
    & Bootstrapping & 93.35  & 92.65 & 91.05 & 80.42 & 81.78 & 82.29  & 80.14 & 73.97 & 64.02 & 80.27 & 77.66 & 72.00 \\
    \cline{2-14}
    & Noisy label regularization & \underline{94.09}  & \underline{93.31} & \underline{92.08} & 80.02 & 81.26 & 84.42  & \underline{84.93} & 79.47 & 67.55 & \underline{82.40} & \underline{80.35} & 75.04 \\ 
    \hline\hline
    \multirow{3}{*}{Correction} 
    & Pixel-wise & 93.82 & 92.81 & 91.42 & 83.68 & 82.75 & 84.41 & 79.18 & 73.58 & 63.29 & 81.36 & 77.89 & 72.32  \\
    \cline{2-14}
    & Adaptive pixel-wise & 93.84 & 93.06 & 92.02 & 78.51 & 79.88 & 82.48 & \textbf{85.36} & \underline{80.02} & \underline{69.33} & 81.79 & 79.94 & \underline{75.32}  \\
    \cline{2-14}
    & AIO2 (proposed) & \textbf{94.58} & \textbf{94.08} & \textbf{93.01} & 84.71 & 83.31 & 81.57  & 82.41 & \textbf{82.17} & \textbf{77.11} & \textbf{83.55} & \textbf{82.74} & \textbf{79.28} \\
    \hline\hline
    \end{tabular}
    \end{center}
    \scriptsize{\hspace{0.5cm} Note: the best and second best results among LNL methods are highlighted in bold and underlined, respectively.}
\end{table*}
\renewcommand{\arraystretch}{1}

\subsection{Experimental Results} \label{sec:exp:results}

\subsubsection{Massachusetts dataset} \label{sec:exp:results:massachussets}
We first show the performance of various compared methods during the training in \cref{fig:exp:hsota}. It can be seen that our proposed AIO2 method achieves the best results no matter how severely the labels are corrupted. In situations with low label noise rates, AIO2 is able to attain results comparable to those when training with GT, while in situations with high label noise rates, AIO2 has a more significant advantage over other LNL methods. Additionally, the use of the ACT module can appropriately trigger the correction program before the model overfits to noisy labels. Note that the model showcases the memorization effects, that is, the performance is first improved and then degraded on all the training data of different label noise rates. It indicates that memorization effects can be generally observed in the learning from noisy labels, thus making it possible for the ACT module to work in common cases. Furthermore, the methods related to label correction strategies, especially those using adaptive thresholds, typically perform better than regularizations.
Bootstrapping can partially reduce the influence of label noise in the training process. However, the consistency constraint has a limited impact, particularly when the labels are highly contaminated. Similarly, noisy label regularization provides only marginal improvements, and in some cases, can even worsen the performance of the pixel-wise label correction method when the quality of training labels is extremely poor.

% seg maps - Boston
\begin{figure*}
    \centering
    \footnotesize
    \begin{tabular}{p{3cm}<{\centering}p{3cm}<{\centering}p{3cm}<{\centering}p{3cm}<{\centering}p{3cm}<{\centering}}
    \includegraphics[width=1.0\linewidth]{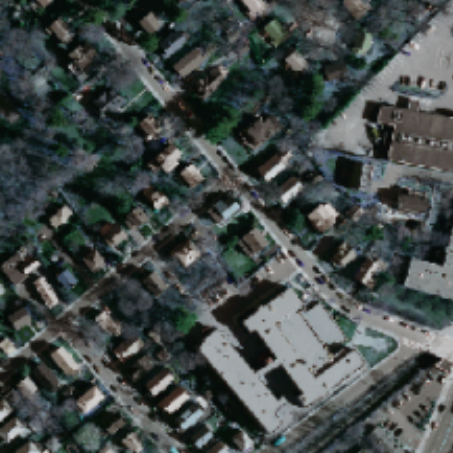} &
    \includegraphics[width=1.0\linewidth]{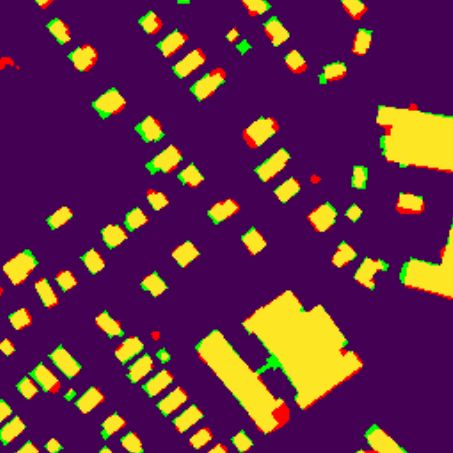} &
    \includegraphics[width=1.0\linewidth]{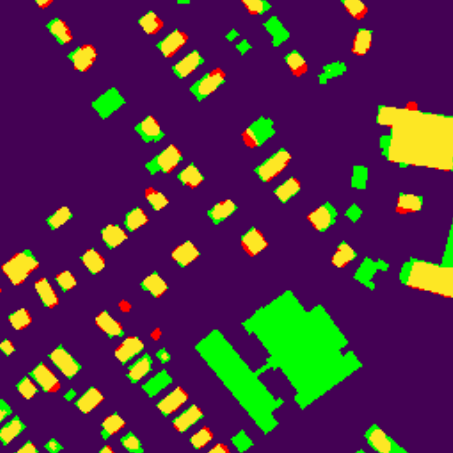} &
    \includegraphics[width=1.0\linewidth]{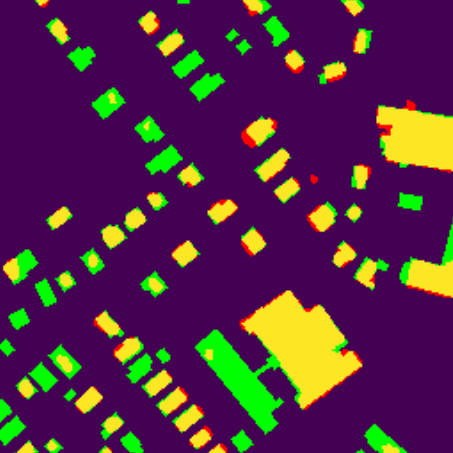} &
    \includegraphics[width=1.0\linewidth]{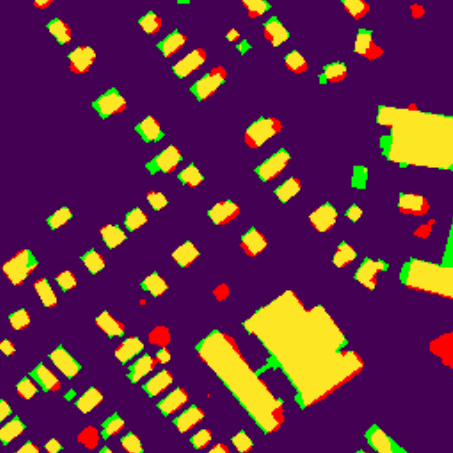} \\
    \end{tabular}
    \begin{tabular}{p{2.5cm}<{\centering}p{3cm}<{\centering}p{3.2cm}<{\centering}p{3.4cm}<{\centering}p{2.4cm}<{\centering}}
    (a) Optical image & \hspace{0.5cm} (b) Train w/ GT & (c) Train w/ noisy labels & (d) Consistency constraint & (e) Bootstrapping \\
    \end{tabular}
    \begin{tabular}{p{3cm}<{\centering}p{3cm}<{\centering}p{3cm}<{\centering}p{3cm}<{\centering}p{3cm}<{\centering}}
    \includegraphics[width=1.0\linewidth]{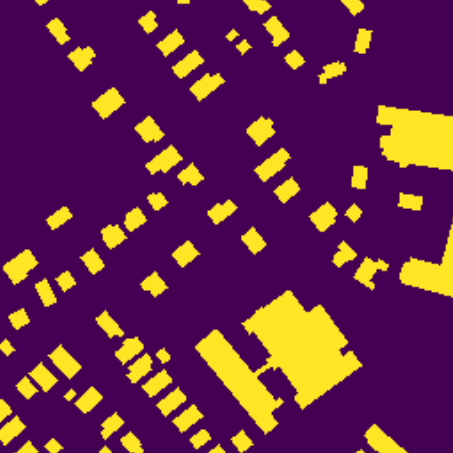} &
    \includegraphics[width=1.0\linewidth]{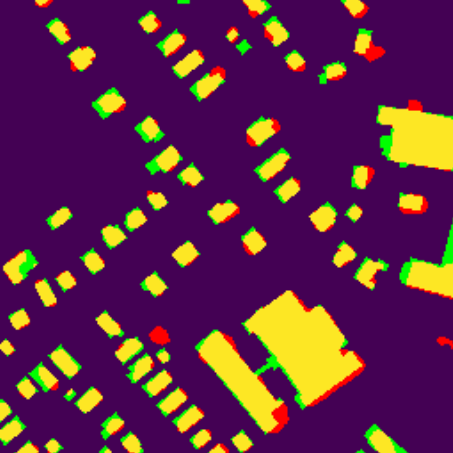} &
    \includegraphics[width=1.0\linewidth]{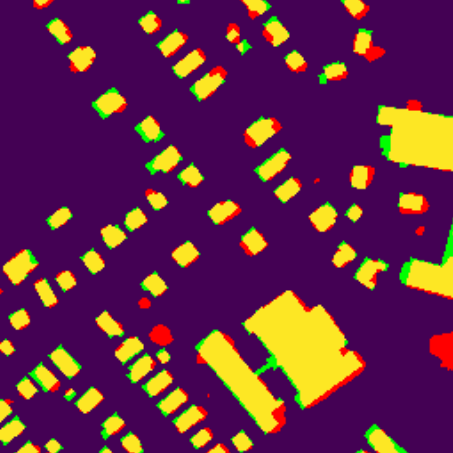} &
    \includegraphics[width=1.0\linewidth]{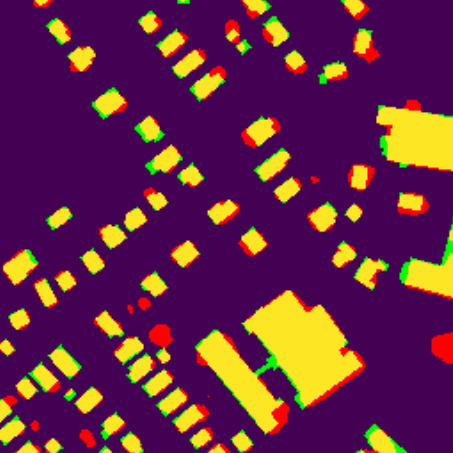} &
    \includegraphics[width=1.0\linewidth]{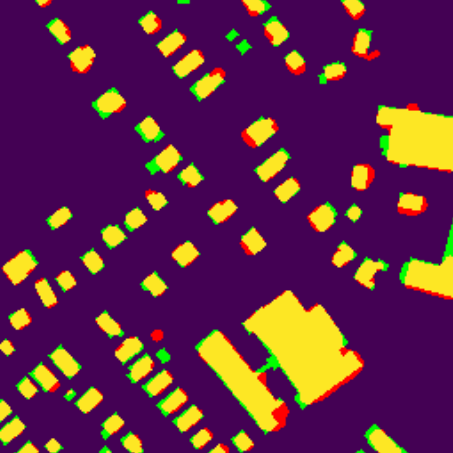} \\
    \end{tabular}
    \begin{tabular}{p{3cm}<{\centering}p{2.5cm}<{\centering}p{3.2cm}<{\centering}p{3.8cm}<{\centering}p{2.8cm}<{\centering}}
    (f) GT & \hspace{0.2cm} (g) Pixel-wise & \hspace{0.07cm} (h) Adaptive pixel-wise & (i) Noisy label regularization & (j) AIO2 (proposed) \\ 
    \end{tabular}
    \caption{{Segmentation maps obtained by considered methods after 300 epoch training on noisy labels with $\alpha_0=0.5$ for the Massachusetts dataset, where false positive and false negative are highlighted with \textcolor{red}{red} and \textcolor{green}{green}, respectively.}}
    \label{fig:exp:hsota-map}
\end{figure*}

Some statistical results are shown in \cref{tab:exp:hsota-iou,tab:exp:hsota-acc}. In general, AIO2 performs the best among all LNL methods, achieving the highest IoUs, OAs, and F1 scores across different levels of label noise. Furthermore, AIO2 exhibits the most stable performance, with the small standard deviations as well as the final results very close to the best ones that the model can ever achieve during training. Moreover, from \cref{tab:exp:hsota-acc} we can observe that the recall is still quite low by consistency constraint, although the precision is improved slightly, indicating that the consistency constraint can help the model learn details, but fails to combat missing instance annotations.

Finally, we present some visual results in \cref{fig:exp:hsota-map}, from which we can draw the same conclusions as before. The proposed AIO2 can generate better segmentation maps than other considered LNL methods, and is able to detect almost all the instances in the scene and also depict the shapes better than others. On the other hand, the model trained purely on noisy labels (Train w/ noisy labels) overfits to incomplete label noise, thereby omitting a number of houses in the segmentation map. While the consistency constraint improves this a bit, some instances are still excluded, as shown in \cref{fig:exp:hsota-map} (d).

\subsubsection{Germany dataset} \label{sec:exp:results:germany}

\begin{figure*}[htp]
    \centering
    \small
    \includegraphics[width=0.95\linewidth]{Figures/Results_sota_legend_1row.pdf}
    \begin{tabular}{p{5.5cm}<{\centering}p{5.5cm}<{\centering}p{5.4cm}<{\centering}}
        \includegraphics[width=1.\linewidth]{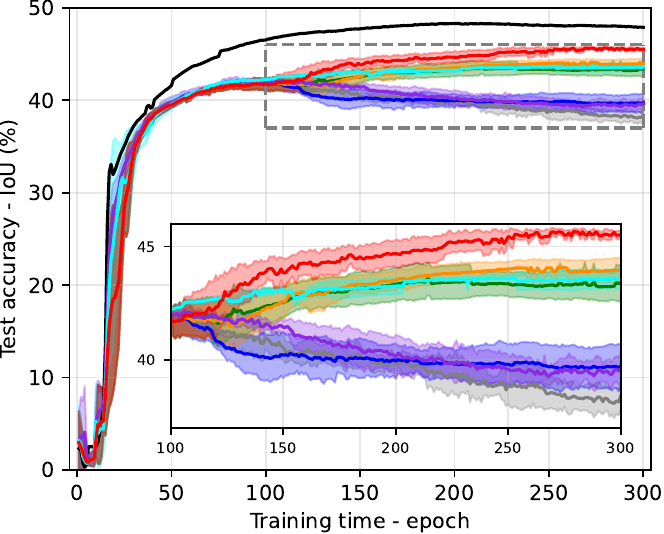} &  
        \includegraphics[width=1.\linewidth]{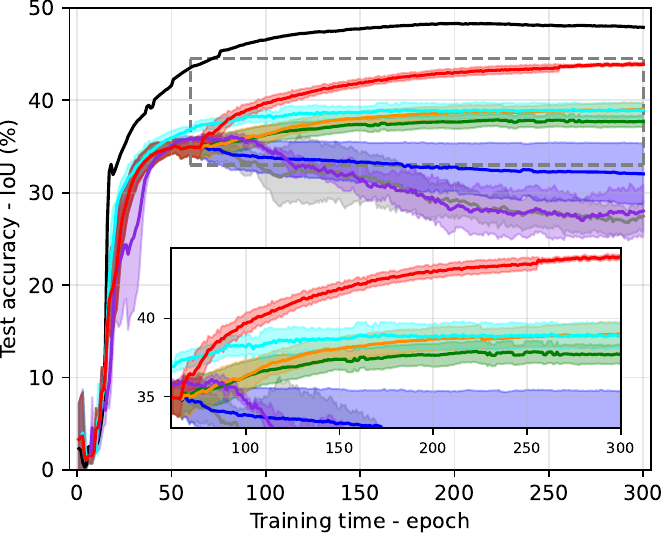} &
        \includegraphics[width=1.\linewidth]{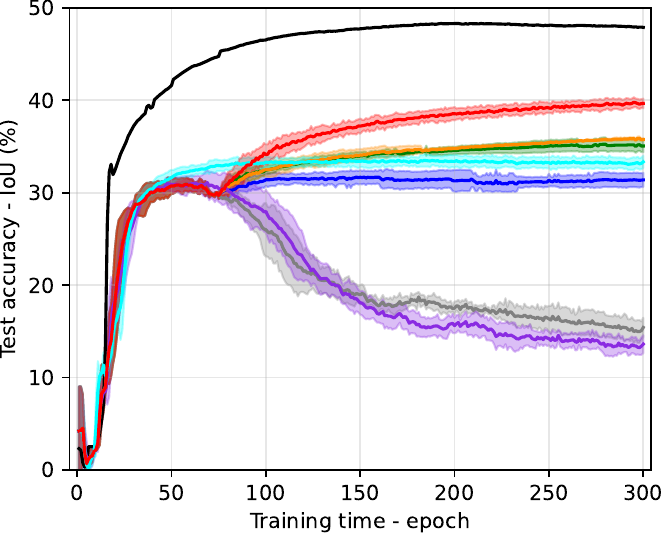} \\
        \hspace{1cm} (a) $\alpha_0$=0.3 & 
        \hspace{1cm} (b) $\alpha_0$=0.5 & 
        \hspace{1cm} (c) $\alpha_0$=0.7
    \end{tabular}
    \caption{Test accuracy (IoU) versus training time (epoch) obtained by considered methods trained with GT or incomplete noisy labels under different given discarding percentages ($\alpha_0$) on the Germany dataset.}
    \label{fig:exp:gsota}
\end{figure*}

\cref{fig:exp:gsota} first illustrates the real-time performance of the model as the training proceeds. AIO2 can promptly trigger the ACT module before the model starts to overfit to label noise, partially leading to the best performance among all compared LNL methods. However, pixel-wise correction based methods do not perform better than bootstrapping to the extent they do on the Massachusetts dataset. This is possibly caused by the lower spatial resolution of planet images, which amplifies the uncertainty of individual pixel samples.

% Germany - comparison result figures

The corresponding IoU statistics and other accuracy results are presented in \cref{tab:exp:gsota-iou,tab:exp:gsota-acc}, respectively. 
The superiority of the proposed AIO2 method is evident from its exceptional performance in terms of IoU, OA, and F1 scores, as well as its enhanced stability. In addition, although pixel-wise correction strategies exhibit higher recall rates than AIO2 on the Massachusetts dataset, AIO2 demonstrates higher precision rates on this Germany dataset. Nevertheless, the harmonic meandofprecision and recall--that is, the F1 scores--suggest that AIO2 still outperforms other compared LNL methods on both datasets. These results highlight the effectiveness of the O2C module at utilizing spatial information to balance the number of samples to correct when applied to different datasets.

% Germany - comparison result IoU tables
\renewcommand{\arraystretch}{1.2}
\begin{table*}[htp]
    \begin{center}
    \scriptsize
    \caption{IoUs obtained by considered methods after 300 epochs on the Germany dataset.}%, where the best and second best results among LNL methods are highlighted in bold and underlined, respectively.}
    \label{tab:exp:gsota-iou}
    \begin{tabular}{c|c?c|c|c?c|c|c}
    \hline\hline
    \multicolumn{2}{c?}{IoU (\%)} & \multicolumn{3}{c?}{Final} & \multicolumn{3}{c}{Maximum} \\
    \hline
    \multicolumn{2}{c?}{$\alpha_0$} & 0.3 & 0.5 & 0.7 & 0.3 & 0.5 & 0.7 \\
    \hline\hline
    \multirow{2}{*}{Baseline} & Train w/ GT & \multicolumn{3}{c?}{48.10} & \multicolumn{3}{c}{48.32} \\
    \cline{2-8}
    & Train w/ noisy labels & 38.42$\pm$0.59 & 27.46$\pm$1.72 & 15.42$\pm$1.05 & 42.03$\pm$0.62 & 35.84$\pm$0.59 & 31.18$\pm$0.65 \\
    \hline\hline
    \multirow{3}{*}{Regularization} 
    & Consistency constraint & 39.60$\pm$0.66 & 28.03$\pm$2.80 & 13.61$\pm$0.97 & 42.38$\pm$0.26 & 36.32$\pm$0.39 & 31.51$\pm$0.82 \\
    \cline{2-8}
    & Bootstrapping & 43.59$\pm$0.46 & 38.96$\pm$0.85 & 33.34$\pm$0.67 & 43.75$\pm$0.25 & \underline{39.15$\pm$0.80} & 33.78$\pm$0.63 \\
    \cline{2-8}
    & Noisy label regularization & \underline{43.97$\pm$0.54} & \underline{39.01$\pm$0.75} & \underline{35.76$\pm$0.14} & \underline{44.13$\pm$0.37} & 39.05$\pm$0.73 & \underline{35.91$\pm$0.20} \\ 
    \hline\hline
    \multirow{3}{*}{Correction} 
    & Pixel-wise & 39.73$\pm$0.95 & 32.03$\pm$3.35 & 31.38$\pm$0.76 & 42.02$\pm$0.62 & 35.64$\pm$0.64 & 31.86$\pm$0.71 \\
    \cline{2-8}
    & Adaptive pixel-wise & 43.40$\pm$0.79 & 37.73$\pm$0.63 & 35.06$\pm$0.67 & 43.65$\pm$0.74 & 37.97$\pm$0.71 & 35.33$\pm$0.69 \\
    \cline{2-8}
    & AIO2 (proposed) & \textbf{{45.52$\pm$0.21}} & \textbf{{43.91$\pm$0.12}} & \textbf{{39.66$\pm$0.52}} & \textbf{{45.80$\pm$0.11}} & \textbf{{43.98$\pm$0.17}} & \textbf{{39.83$\pm$0.49}} \\
    \hline\hline
    \end{tabular}
    \end{center}
    \scriptsize{\hspace{0.8cm} Note: the best and second best results among LNL methods are highlighted in bold and underlined, respectively.}
\end{table*}
\renewcommand{\arraystretch}{1}

% Germany - comparison result accuracy tables
\renewcommand{\arraystretch}{1.2}
\begin{table*}[htp]
    \begin{center}
    \scriptsize
    \caption{OAs, precisions, recalls, and F1 scores obtained by considered methods after 300 epochs on the Germany dataset.}%, where the best and second best results among LNL methods are highlighted in bold and underlined, respectively.}
    \label{tab:exp:gsota-acc}
    \begin{tabular}{c|c?c|c|c?c|c|c?c|c|c?c|c|c}
    \hline\hline
    \multicolumn{2}{c?}{Accuracy (\%)} & \multicolumn{3}{c?}{OA} & \multicolumn{3}{c?}{Precision} & \multicolumn{3}{c?}{Recall} & \multicolumn{3}{c}{F1}\\
    \hline
    \multicolumn{2}{c?}{$\alpha_0$} & 0.3 & 0.5 & 0.7 & 0.3 & 0.5 & 0.7 & 0.3 & 0.5 & 0.7 & 0.3 & 0.5 & 0.7 \\
    \hline\hline
    \multirow{2}{*}{Baseline} & Train w/ GT & \multicolumn{3}{c?}{93.37} & \multicolumn{3}{c?}{69.48} & \multicolumn{3}{c?}{61.36} & \multicolumn{3}{c}{65.18} \\
    \cline{2-14}
    & Train w/ noisy labels & 92.74 & 91.99 & 90.70 & \underline{71.84} & \underline{76.97} & \underline{77.66} & 45.87 & 31.45 & 16.93 & 55.98 & 44.60 & 27.78 \\
    \hline\hline
    \multirow{3}{*}{Regularization} 
    & Consistency constraint & \underline{92.83} & 92.05 & 90.57 & \textbf{72.94} & \textbf{78.17} & \textbf{81.75} & 46.95 & 32.30 & 14.77 & 57.11 & 45.58 & 24.99 \\
    \cline{2-14}
    & Bootstrapping & 92.60 & \underline{92.53} & \underline{92.13} & 63.13 & 64.85 & 65.45 & 58.49 & 49.87 & 41.12 & 60.70 & 56.36 & 50.49 \\
    \cline{2-14}
    & Noisy label regularization & 92.35 & 91.54 & 91.77 & 59.51 & 53.86 & 57.96 & \underline{62.91} & \underline{60.35} & \underline{49.81} & \underline{61.09} & \underline{56.89} & \underline{53.51} \\ 
    \hline\hline
    \multirow{3}{*}{Correction} 
    & Pixel-wise & 92.57 & 91.41 & 91.37 & 66.69 & 57.25 & 58.78 & 50.04 & 46.32 &  43.18 & 56.97 & 50.04 & 49.56 \\
    \cline{2-14}
    & Adaptive pixel-wise & 91.98 & 91.06 & 91.14 & 57.70 & 50.90  & 53.84 & \textbf{64.36} & \textbf{61.16} & \textbf{52.31} & 60.75 & 56.89 & 53.51\\
    \cline{2-14}
    & AIO2 (proposed) & \textbf{{93.22}} & \textbf{{93.07}} & \textbf{{92.59}} & {67.68} & {65.58} & {65.45} & {58.34}  & {57.19} & {50.52} & \textbf{{62.65}} & \textbf{{61.09}} & \textbf{{57.02}} \\
    \hline\hline
    \end{tabular}
    \end{center}
    \scriptsize{\hspace{0.5cm} Note: the best and second best results among LNL methods are highlighted in bold and underlined, respectively.}
\end{table*}
\renewcommand{\arraystretch}{1}

\begin{figure*}
    \centering
    \small
    \begin{tabular}{p{3cm}<{\centering}p{3cm}<{\centering}p{3cm}<{\centering}p{3cm}<{\centering}p{3cm}<{\centering}}
    \includegraphics[width=1.0\linewidth]{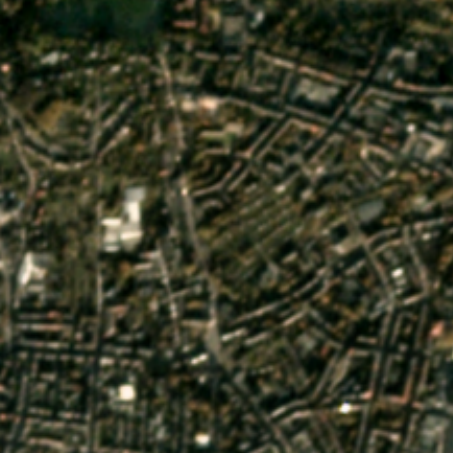} &
    \includegraphics[width=1.0\linewidth]{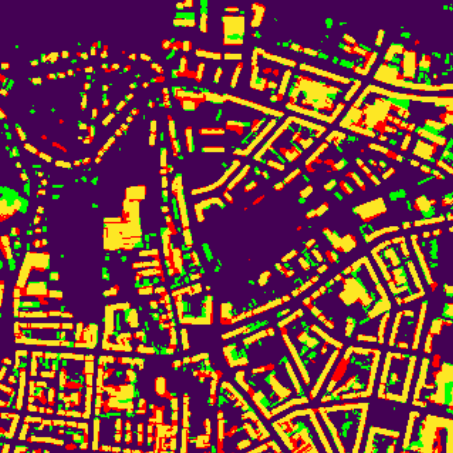} &
    \includegraphics[width=1.0\linewidth]{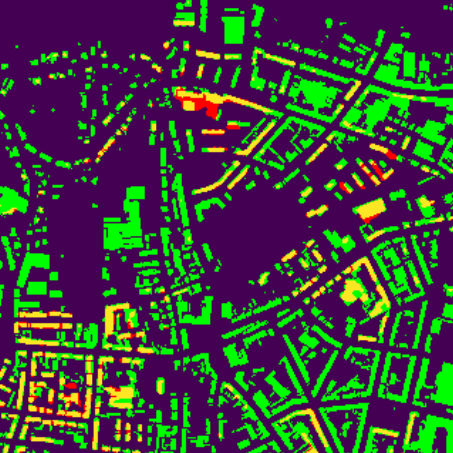} &
    \includegraphics[width=1.0\linewidth]{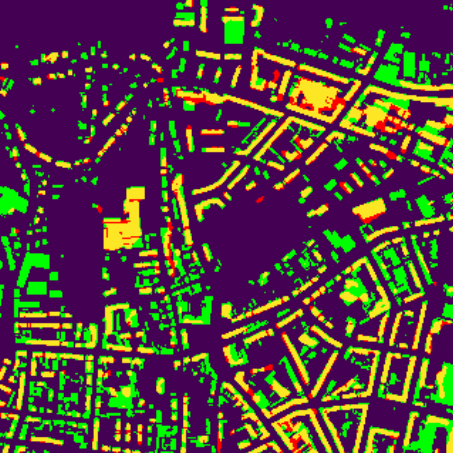} &
    \includegraphics[width=1.0\linewidth]{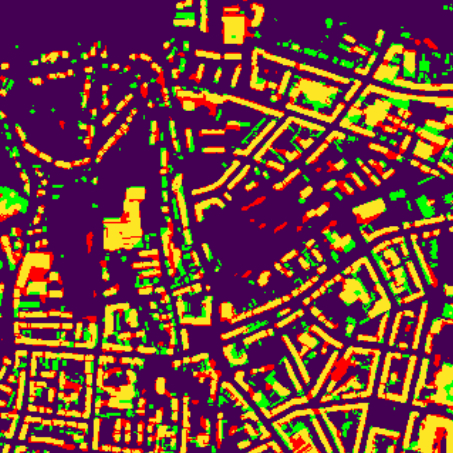} \\
    \end{tabular}
    \begin{tabular}{p{2.5cm}<{\centering}p{3cm}<{\centering}p{3.2cm}<{\centering}p{3.4cm}<{\centering}p{2.4cm}<{\centering}}
    (a) Optical image & \hspace{0.5cm} (b) Train w/ GT & (c) Train w/ noisy labels & (d) Consistency constraint & (e) Bootstrapping \\
    \end{tabular}
    \begin{tabular}{p{3cm}<{\centering}p{3cm}<{\centering}p{3cm}<{\centering}p{3cm}<{\centering}p{3cm}<{\centering}}
    \includegraphics[width=1.0\linewidth]{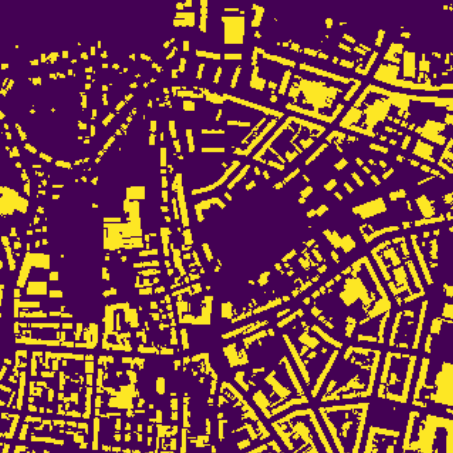} &
    \includegraphics[width=1.0\linewidth]{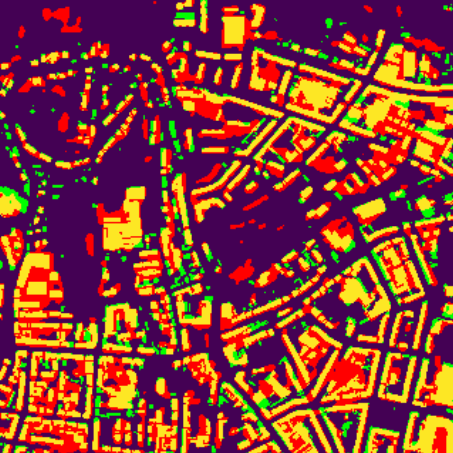} &
    \includegraphics[width=1.0\linewidth]{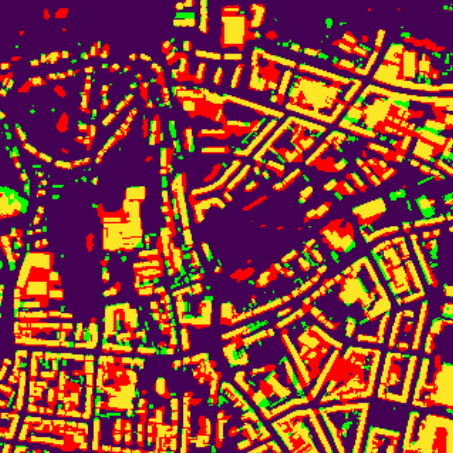} &
    \includegraphics[width=1.0\linewidth]{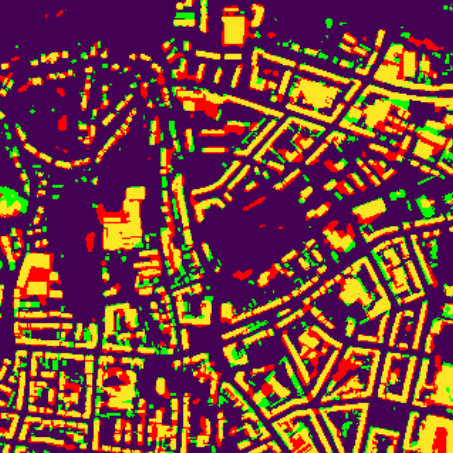} &
    \includegraphics[width=1.0\linewidth]{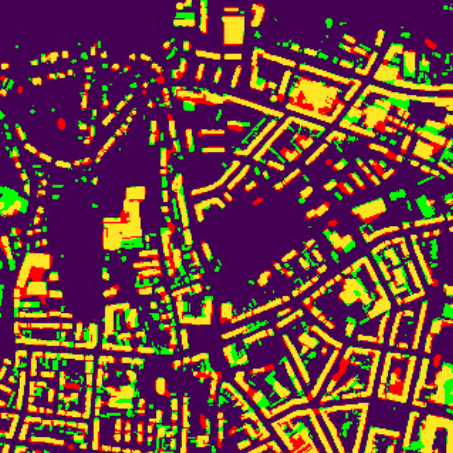} \\
    \end{tabular}
    \begin{tabular}{p{3cm}<{\centering}p{2.5cm}<{\centering}p{3.2cm}<{\centering}p{3.8cm}<{\centering}p{2.8cm}<{\centering}}
    (f) GT & \hspace{0.2cm} (g) Pixel-wise & \hspace{0.07cm} (h) Adaptive pixel-wise & (i) Noisy label regularization & (j) AIO2 (proposed)  \\ 
    \end{tabular}
    \caption{{Segmentation maps obtained by considered methods after 300 epoch training on noisy labels with $\alpha_0=0.5$ for the Germany dataset, where false positive and false negative are highlighted with \textcolor{red}{red} and \textcolor{green}{green}, respectively.}}
    \label{fig:exp:gsota-map}
\end{figure*}

To visually evaluate the effectiveness of the proposed method, \cref{fig:exp:gsota-map} presents a series of segmentation maps for a specific test image by various compared methods. These segmentation maps demonstrate that AIO2 produces the top-performing outcome among all LNL methods, with a lower false positive rate and a clearer portrayal of details, followed by bootstrapping. {However, there tends to be more green parts, that is, the false negative parts, in the map by AIO2 than those by the noisy label regularization method, when compared to the results on the Massachusetts dataset. We attribute this phenomenon to the lower spatial resolution of planet images. Many predicted objects tend to be connected to each other, thereby discarded by the overlap check in O2C.} Pixel-wise correction methods typically result in over-segmentation maps. This problem can be partially alleviated by employing noisy label regularization. In contrast, the maps generated by training with noisy labels and the consistency constraint are clearly impaired by the incompleteness issue as a consequence of overfitting to noisy labels. 

\subsection{Ablation Studies} \label{sec:exp:ablation}

In addition to the experiments discussed above, we conducted ablation studies to test the roles of the newly designed ACT module, the teacher model, and the soft boundary trick in the O2C module. 

\subsubsection{Adaptive correction trigger (ACT) module}
\begin{figure*}[htp]
    \centering
    \begin{tabular}{p{4cm}<{\centering}p{4cm}<{\centering}p{4cm}<{\centering}p{4cm}<{\centering}}
        \includegraphics[width=1.07\linewidth]{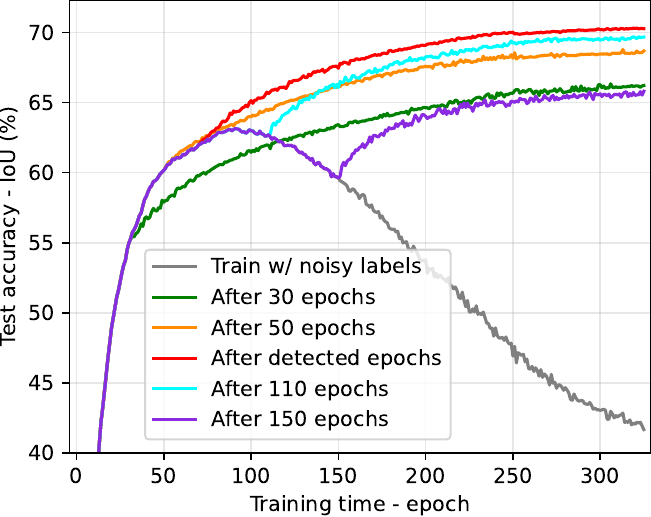} &  
        \includegraphics[width=1.09\linewidth]{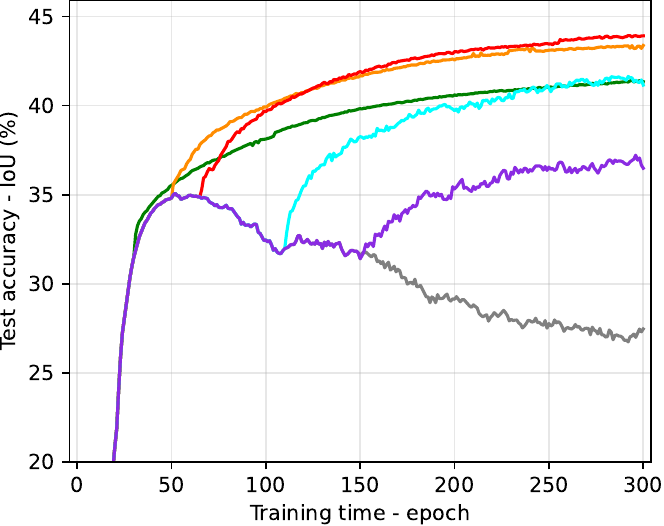} &  
        \includegraphics[width=1.07\linewidth]{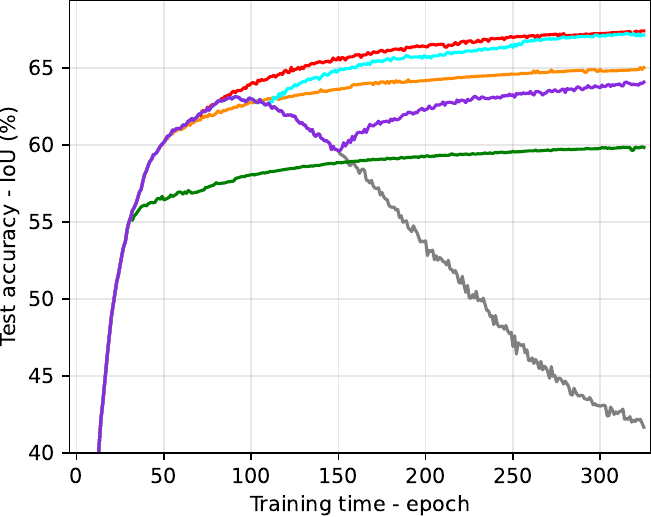} &
        \includegraphics[width=1.09\linewidth]{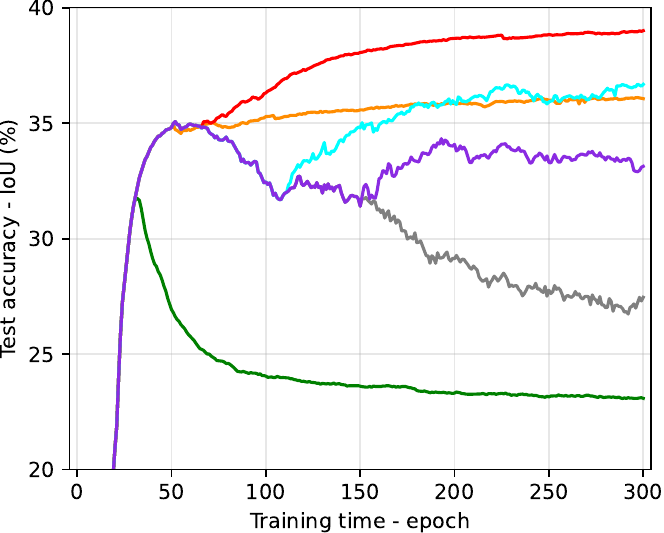}  \\
    \end{tabular}
    \begin{tabular}{p{3.7cm}<{\centering}p{3.5cm}<{\centering}p{4.5cm}<{\centering}p{3.5cm}<{\centering}}
        \hspace{0.5cm} \small (a) O2C+Massachusetts & 
        \hspace{0.9cm} \small (b) O2C+Germany & 
        \hspace{0.4cm} \small (c) Pixel-wise+Massachusetts & 
        \hspace{0.2cm} \small (d) Pixel-wise+Germany 
    \end{tabular}
    \caption{Test accuracy (IoU) versus training time (epoch) obtained by combining object-wise (O2C) and pixel-wise label correction strategies with different numbers of warm-up epochs on the Massachusetts and Germany datasets, of which the labels are corrupted with a discarding percentage $\alpha_0=0.5$. Specifically, the results wrt pixel-wise correction were generated by class-wise adaptive threshold with noisy label regularization.}
    \label{fig:exp:el}
\end{figure*}

To evaluate the necessity of using ACT in label correction methods, we conducted tests on both object-wise and pixel-wise label correction strategies triggered at different numbers of epochs. The results are presented in \cref{fig:exp:el}. These results suggest that the timing of the label correction process has a significant impact on the final performance. Starting the label correction procedure too early (when the model is still underfitting) or too late (when the model starts to overfit to noisy labels) can both lead to suboptimal results. In this context, the proposed ACT can effectively mitigate these negative effects by replacing the manual warm-up stage setting with adaptive early learning detection.

% early-learning figures

Additionally, \cref{tab:exp:el} presents detailed results of early learning detection by ACT. It displays the model's performance on the training set at detected epochs in comparison with the best counterpart achieved by models  when directly trained with noisy labels. Two pieces of information can be gleaned from this table. First, the detected models are very close to the best-performing ones, which partly explains why ACT can help models obtain the promising results shown in \cref{fig:exp:el}. Second, the repeated implementation of ACT shows stable detection performance. {Note that while on the Massachusetts dataset with $\alpha_0=0.3$ there is little difference in model performance between the detected and the best models, the ACT module can still guarantee the quality of pseudo labels in the O2C module (see \cref{fig:exp:hsota}). This is partly due to the high resolution of images (lower pixel uncertainties) and a less severe corruption of labels. In this case, models directly trained on noisy labels would experience less significant damage, leading to an elongated and flattened transition phase where the peak performance deviates from the center point.}

\subsubsection{Teacher model} \label{sec:exp:ablation:teacher}

\begin{figure}[htp]
    \centering
    \begin{tabular}{p{4cm}<{\centering}p{4cm}<{\centering}}
        \includegraphics[width=1.03\linewidth]{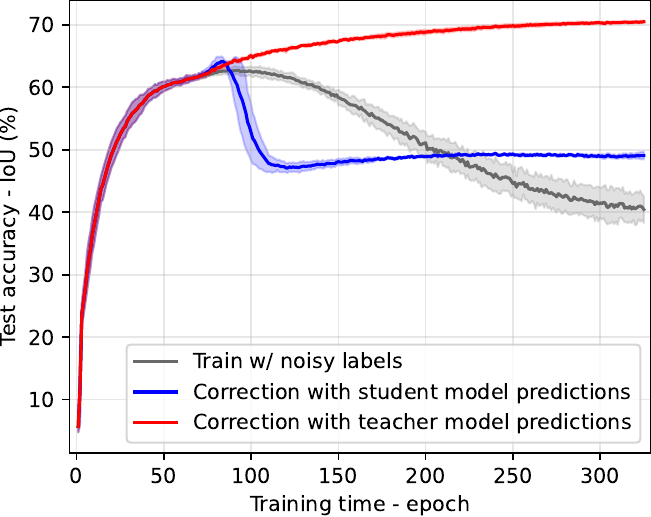} & 
        \includegraphics[width=1.03\linewidth]{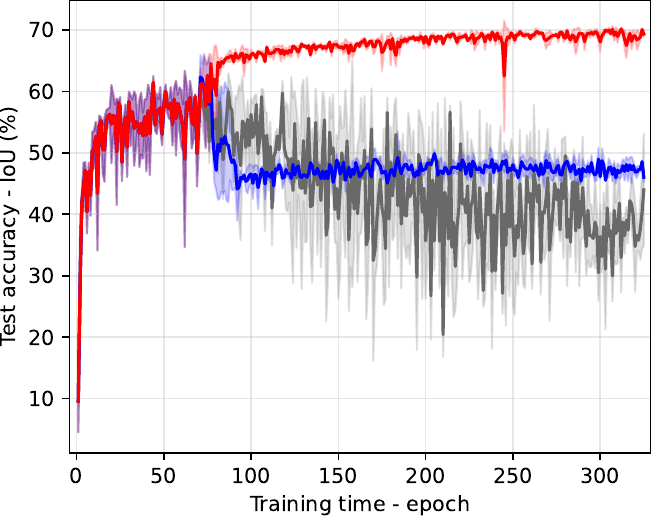} \\
        \hspace{0.5cm} \small (a) Teacher model & 
        \hspace{0.5cm} \small (b) Student model
    \end{tabular}
    \caption{Test accuracies of (a) teacher models and (b) student models by using teacher (red) or student (blue) model predictions as pseudo labels in O2C on the Massachusetts dataset with noisy labels generated under $\alpha_0=0.5$.}
    \label{fig:exp:sobj}
\end{figure}

As mentioned in \cref{sec:meth:o2c}, we decouple model training from the label correction process by utilizing the predictions of teacher models as pseudo labels. By way of comparison, we employ student models as a substitute for the teacher models to act as the corrected label source in the O2C module. \Cref{fig:exp:sobj} illustrates the results of two cases, demonstrating that model training collapses to some degree when label correction is initiated with student model predictions as pseudo labels.

\begin{table*}[htp]
    \begin{center}
    \caption{Early-learning detection results of the numbers of epochs where the label correction is triggered by the ACT module.}% along with corresponding training accuracies wrt GT (accuracies of the detected epoch/maximum during the training in brackets).}
    \label{tab:exp:el}
    \begin{tabular}{c|c|ccc|c}
        \hline\hline
        %  & \multirow{2}{*}{$\alpha_0$} & \multicolumn{3}{c|}{Replay No.} & \multirow{2}{*}{avg.}\\
        %  &  & 1 & 2 & 3 & \\
        & $\alpha_0$ & Replay 1 & Replay 2 & Replay 3 & avg. \\
        \hline\hline
        \multicolumn{1}{c|}{\multirow{3}{*}{Massachusetts}} 
        & 0.3 & 93 (67.66/70.36) & 89 (67.32/70.67) & 101 (67.08/70.31)  & 94 (67.35/70.45) \\ 
        & 0.5 & 82 (62.39/63.04) & 70 (62.11/62.90) & 73 (61.79/62.76)   & 75 (62.10/62.90) \\ 
        & 0.7 & 68 (52.10/53.02) & 80 (53.08/53.33) & 87 (54.24/54.37)   & 78 (53.14/53.58) \\
        \hline\hline
        \multicolumn{1}{c|}{\multirow{3}{*}{Germany}} 
        & 0.3 & 115 (41.22/41.68) & 107 (41.93/41.93) & 119 (41.32/41.61) & 114 (41.49/41.74)\\ 
        & 0.5 & 67 (35.23/35.41)   & 76 (32.90/33.38)   & 67 (33.64/33.82)   & 70 (33.92/34.20)\\ 
        & 0.7 & 77 (26.26/26.50)   & 73 (26.16/26.51)   & 80 (25.71/26.87)   & 77 (26.04/26.63) \\
        \hline\hline
    \end{tabular}
    \end{center}
    \scriptsize{\hspace{2.6cm} Note: the training accuracies wrt GT (those at the detected epoch/maximum during the training) are shown in brackets.}
\end{table*}

% student model for correction

\begin{figure}[htp]
    \centering
    \begin{tabular}{p{3.8cm}<{\centering}p{3.8cm}<{\centering}}
        \includegraphics[width=1.07\linewidth]{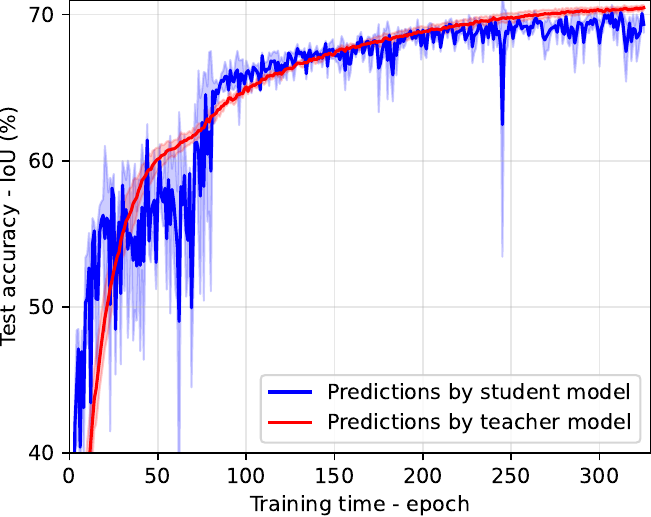} &
        \includegraphics[width=1.09\linewidth]{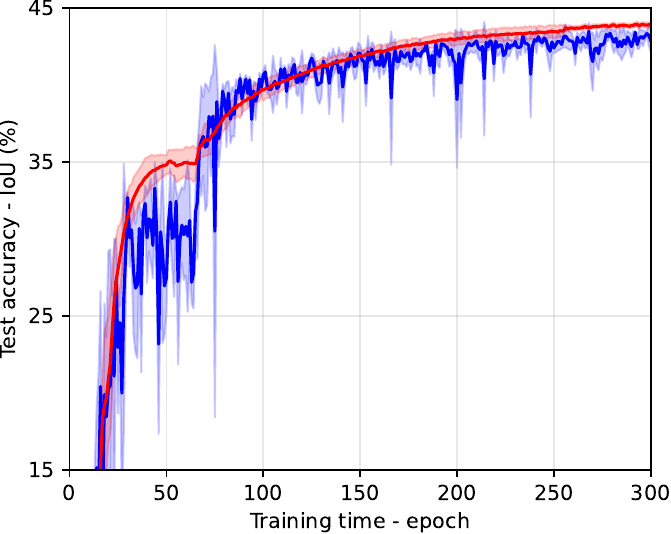}  \\
        \hspace{0.5cm} \small (a) Massachusetts & 
        \hspace{0.5cm} \small (b) Germany 
    \end{tabular}
    \caption{Test accuracies of teacher (red) and student (blue) models trained by the proposed AIO2 on the (a) Massachusetts and (b) Germany datasets with noisy labels generated under $\alpha_0=0.5$.}
    \label{fig:exp:st}
\end{figure}

In addition, we plot test accuracies obtained by teacher and student models in AIO2 as a function of training time, as shown in \cref{fig:exp:st}. Clearly, teacher models can achieve both better and more stable results than student models, thereby enhancing the overall robustness of the proposed method.

% role of teacher model

\subsubsection{Soft boundary trick} \label{sec:exp:ablation:softb}
Rather than using hard labels, O2C applies a soft filter on correction candidates in order to generate soft labels around boundaries. The reason behind this is that the boundary samples naturally enjoy a higher level of uncertainty and are more likely to be misclassified. We conduct experiments on two datasets by removing the soft boundary trick and trying different filter sizes. The results shown in {\cref{tab:exp:fs}} demonstrate the effectiveness of the soft boundary trick, which improves the model performance by approximately 2 percentage points on both datasets. Moreover, the model is not overly sensitive to the filter size, except for a slight drop in performance when using a filter size of 7 on the Germany dataset. This is reasonable, since the spatial resolution of this dataset (3m) is relatively low. A $7\times7$ filter covers an area of roughly 441m$^2$, which is big enough to mix up everything for building extraction. This is also why we chose a smaller soft filter size (3) empirically for the Germany dataset than for the Masschusetts dataset. {Thus, we take the filter size as a fine-tuning hyperparameter, and recommend that readers empirically adjust the filter size based on the target application and the spatial resolution of images. For instance, segmentation of large objects such as urban green spaces, or a higher spatial resolution (e.g., better than 1m) would be better served by a bigger filter size, while a smaller filter size better suits tiny objects like roads, or a lower spatial resolution. An unrealistic filter size is probably harmful to model performance.}

% soft boundary trick
\begin{table}[]
    \begin{center}
    \caption{Final test accuracies obtained by AIO2 using different soft filter sizes on the Massachusetts and Germany datasets with noisy labels generated under $\alpha_0=0.5$.}%, where the results by default settings for each dataset in the previous sections are highlighted in bold.}
    \label{tab:exp:fs}
    \begin{tabular}{c|c|c}
        \hline\hline
         Filter size & Massachusetts  & Germany \\
         \hline\hline
         0           & 68.67$\pm$0.25 & {42.02$\pm$0.89}\\
        %  \hline
         3           & 70.29$\pm$0.26 & \textbf{{43.91$\pm$0.12}}\\
        %  \hline
         5           & \textbf{70.26$\pm$0.24} & {42.93$\pm$0.62}\\
        %  \hline
         7           & 70.31$\pm$0.32 & {40.35$\pm$0.76}\\
         \hline\hline
    \end{tabular}
    \end{center}
    \scriptsize{Note: the results by default settings for each dataset in the previous sections are highlighted in bold.}
\end{table}

\subsection{Parameter Sensitivity Analysis} \label{sec:exp:param}

\begin{figure*}
    \centering
    \begin{tabular}{p{4cm}<{\centering}p{4cm}<{\centering}p{4cm}<{\centering}p{4cm}<{\centering}}
        \includegraphics[width=1.07\linewidth]{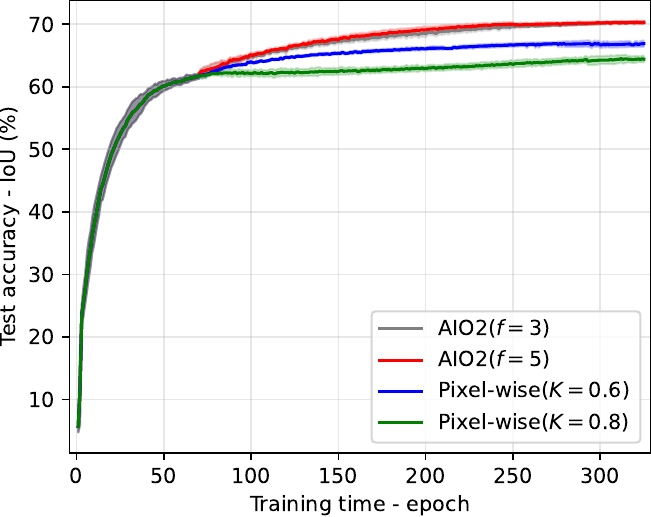} & 
        \includegraphics[width=1.07\linewidth]{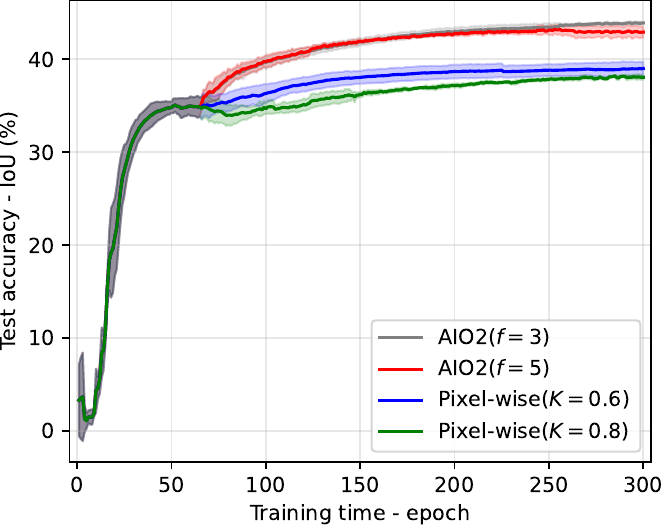} & 
        \includegraphics[width=1.07\linewidth]{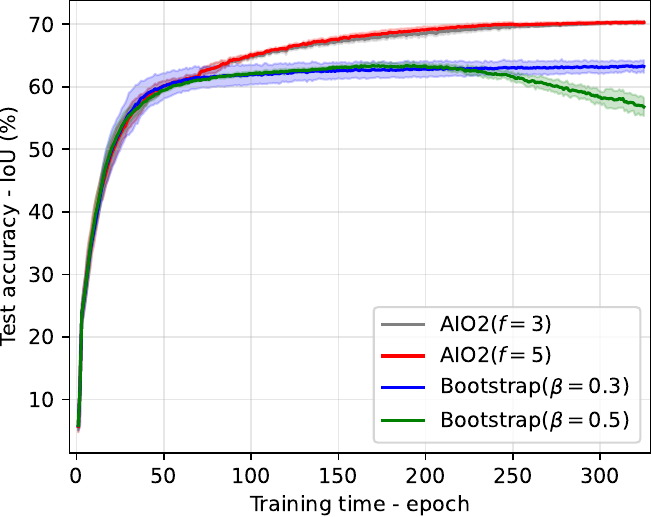} & 
        \includegraphics[width=1.07\linewidth]{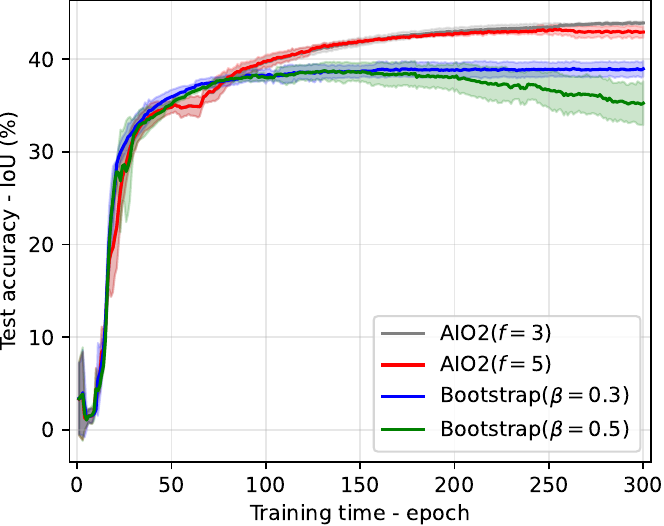} \\
    \end{tabular}
    \begin{tabular}{cccc}%{p{4.5cm}<{\centering}p{4.2cm}<{\centering}p{4.8cm}<{\centering}p{4.0cm}<{\centering}}
        \small (a) v.s. pixel-wise+Massachusetts & 
        \small (b) v.s. pixel-wise+Germany & 
        \small (c) v.s. bootstrapping+Massachusetts & 
        \small (d) v.s. bootstrapping+Germany 
    \end{tabular}
    \caption{Parameter sensitivity analysis in comparison with noisy label regularization and bootstrapping methods on Massachusetts and Germany datasets with noisy labels generated under $\alpha_0 = 0.5$.}
    \label{fig:exp:param}
\end{figure*}

In this section, we compare the parameter sensitivity of the proposed AIO2 with that of noisy label regularization and bootstrapping. Recall that noisy label regularization is based on adaptive pixel-wise label correction using a predefined confidence threshold $K$, while bootstrapping requires a manually set weight $\beta$ between noisy labels and predictions. Therefore, we plot the results with different filter sizes $f$ of AIO2 in \cref{fig:exp:param}, in contrast to those obtained by noisy label regularization with different $K$ and by bootstrapping with different $\beta$. Similar to the results presented in {\cref{tab:exp:fs}}, AIO2 with different values of $f$ performs on a par with each another, whereas changes in $K$ and $\beta$ can significantly affect the performance of their corresponding methods. In fact, an inappropriate setting of $\beta$ in bootstrapping can even lead to a drop in accuracy. These findings demonstrate that the proposed method is less sensitive to parameter settings, making it a promising choice for practical applications.

\section{Conclusions and Perspectives} \label{sec:conclude}

In this work, we introduced and evaluated a novel mechanism to efficiently train binary semantic segmentation models on incomplete labels by means of Adaptively trIggering Online Object-wise correction (AIO2). AIO2 is a fully automatic, iterative label correction framework comprising two key components: the Adaptive Correction Trigger (ACT) module and the Online Object-wise label Correction (O2C) module. Both modules interact without explicitly setting a predefined warm-up phase. While ACT exploits the characteristics of the training accuracy curve over training epochs, O2C features an object-level correction strategy instead of the widely-used pixel-level algorithms. This way, AI02 automates the addition of pseudo labels to the training dataset, and exploits spatial information to assist with sample correction for segmentation. Besides, O2C operates \textit{on-line} with little extra storage required due to the exploitation of a mean teacher model where the exponential moving average also partially decouples the label correction process from the student model training.

Experimental results obtained on two geographically distinct datasets (Germany and the United States) with spatial resolutions varying by about one order of magnitude indicate the effectiveness of the proposed method. For example, when dropping about 30\% of building labels in 1m-resolution overhead imagery, AIO2 yields accuracy improvements of about 10 percentage points compared to naive supervised training with noisy labels. When the spatial resolution decreases by an order of magnitude for the Germany dataset, we still observe improvements of about 5 percentage points showcasing the robustness of AIO2.
{However, we can still observe a larger IoU gap between AIO2 and training with GT labels on the Germany dataset than that on the Massachusetts dataset, which indicates the limitation of the proposed AIO2 method when coping with RS images of a lower resolution.}

{This work is our initial step toward a systematic solution for training deep neural network models from noisy labels for geospatial semantic segmentation. 
In future works, we will devote ourselves to improving the effectiveness of AIO2 on handling contiguous objects on the low-resolution RS images. Furthermore, we will test AIO2 on the combined noisy labels additionally with shape label noise, and expand AIO2's application to multi-class segmentation tasks such as land cover mapping.
Exploring the potential of AIO2 in a multi-round fashion is another interesting topic to investigate.}

\section*{Acknowledgement}
{The work of C. Liu, Y. Wang and C. Albrecht} was funded by the Helmholtz Association through the Framework of \textit{HelmholtzAI}, grant ID: \texttt{ZT-I-PF-5-01} -- \textit{Local Unit Munich Unit @Aeronautics, Space and Transport (MASTr)}. The compute related to this work was supported by the Helmholtz Association's Initiative and Networking Fund on the HAICORE@FZJ partition. {The work of Q. Li and X. Zhu is jointly supported by the Excellence Strategy of the Federal Government and the Länder through the TUM Innovation Network EarthCare  and by the German Federal Ministry of Education and Research (BMBF) in the framework of the international future AI lab "AI4EO -- Artificial Intelligence for Earth Observation: Reasoning, Uncertainties, Ethics and Beyond" (grant number: 01DD20001).} {The authors thank Nikolai Skuppin for sharing the Germany dataset \cite {Skuppin2022igarss} with them, and Nassim AIT ALI BRAHAM for discussions during the group meetings.}

{\small
\bibliographystyle{IEEEtran}
\bibliography{refs}
}

\end{document}